\documentclass[10pt]{article}
\usepackage[preprint]{tmlr}

\usepackage{amsmath,amsfonts,bm}

\def\eqref#1{equation~\ref{#1}}

\def\1{\bm{1}}

\DeclareMathAlphabet{\mathsfit}{\encodingdefault}{\sfdefault}{m}{sl}
\SetMathAlphabet{\mathsfit}{bold}{\encodingdefault}{\sfdefault}{bx}{n}

\usepackage{hyperref}
\usepackage{url}
\usepackage{booktabs}
\usepackage{amsmath,amssymb,amsthm}
\usepackage{graphicx}
\usepackage{xcolor}
\usepackage{algorithm}
\usepackage{algpseudocode}
\usepackage{subcaption}
\usepackage[capitalize,noabbrev]{cleveref}

\newtheorem{proposition}{Proposition}
\newtheorem{definition}{Definition}
\newtheorem{example}{Example}

\theoremstyle{remark}

\newcommand{\X}{\mathcal{X}}
\newcommand{\U}{\mathrm{U}}
\newcommand{\Vol}{V}
\newcommand{\relu}{\operatorname{relu}}

\newcommand{\nRunsMain}{1,320}
\newcommand{\vFedAvgsynthetic}{12}
\newcommand{\vFedAvgadult}{26}
\newcommand{\vFedAvgcompas}{87}
\newcommand{\vFedAvggerman}{7}
\newcommand{\vFedAvgheart}{50}
\newcommand{\vFedAvgacsincome}{26}
\newcommand{\nCells}{30}
\newcommand{\nBoxBeatsData}{30}
\newcommand{\nCegisLeqBox}{30}
\newcommand{\nEquiv}{29}
\newcommand{\nEquivTot}{30}
\newcommand{\abBoxVsDataTwoUTen}{34/40}
\newcommand{\abBoxVsDataTwoPTen}{21/40}
\newcommand{\abBoxOnlyVsDataTwoUTen}{36/40}
\newcommand{\abBoxOnlyVsDataTwoPTen}{13/40}
\newcommand{\abDataTwoVsDataUTen}{19/40}

\newcommand{\abBoxVsDataTwoUHundred}{33/40}
\newcommand{\abBoxVsDataTwoPHundred}{18/40}
\newcommand{\abBoxOnlyVsDataTwoUHundred}{34/40}
\newcommand{\abBoxOnlyVsDataTwoPHundred}{13/40}
\newcommand{\abDataTwoVsDataUHundred}{16/40}

\newcommand{\cenBoxLtData}{28/40}

\title{Monotonicity Under Federated Aggregation:\\ Measuring, Enforcing and Auditing\\ Domain Constraints}

\author{\name Md.\ Ashraful Kabir Alif \\
      \addr BRAC University, Dhaka, Bangladesh \\
      {\small Contact: \href{mailto:md.ashraful.kabir.alif@g.bracu.ac.bd}{\normalfont\small\texttt{md.ashraful.kabir.alif@g.bracu.ac.bd}}}
      \AND
      \name Sadia Afrin \\
      \addr BRAC University, Dhaka, Bangladesh}

\begin{document}

\maketitle

\begin{abstract}
Models for credit, health and criminal-justice decisions are often required to move in a known direction with key features: a larger loan should not raise the chance of a ``good'' credit rating, and more prior offences should not lower predicted risk.
We study how such monotonicity rules fare under federated learning (FL), where no party can pool the data, and how to enforce and audit them.
In \nRunsMain{} runs on six tabular tasks, two of them with natural client splits (hospitals and US states), unconstrained FedAvg models violate standard domain rules on 7--87\% of the input domain.
Because monotonicity is a property of the function rather than of the data, a penalty can be evaluated on points that no client owns.
On the data distribution, gradient penalties on client examples and on samples from the shared input box perform comparably.
Over the whole domain, the box samples leave fewer violations at every penalty strength on every task (\nBoxBeatsData{} of \nCells{} settings) at equal AUC, and ablations show that the location of the points drives this gain rather than their number.
A counterexample-guided variant, CEGIS-FL, keeps the violated volume below 0.2\% per model at 95\% confidence.
Penalties make violations rare but do not remove them: in our experiments only monotone-by-construction networks, which FedAvg preserves exactly, could be certified, and they matched unconstrained accuracy.
We also show that a widely used random-pair violation rate is not a monotonicity measure.
Code, data pipelines and run logs are released.
\end{abstract}

\section{Introduction}
\label{sec:intro}

Many decision models are expected to respect \emph{directional} domain knowledge.
Under a monotonicity rule, increasing a feature such as the number of prior offences, the duration of a loan or exercise-induced ST depression must never move the prediction in the wrong direction, whatever the values of the other features.
Such rules make models easier to audit and to justify, and they rule out decisions that are hard to defend, such as a credit model that rewards a larger loan.
Monotone models have a long history, from monotonicity hints \citep{sill1996hints} and monotonic networks \citep{sill1997monotonic} to lattice models \citep{gupta2016lattice,you2017dln} and certified or constrained neural networks \citep{liu2020certified,sivaraman2020comet,runje2023cmnn,nolte2023expressive}.
All of this work assumes that the training data can be pooled.

Federated learning (FL; \citealp{mcmahan2017fedavg,kairouz2021advances}) removes that assumption, and makes monotonicity harder to obtain and to check.
Clients see different parts of the feature space (label skew, covariate skew, different hospitals or regions), so local data can show trends that contradict the global rule.
Meanwhile, the server, which is the natural place to check the model, never sees an input.
A tempting shortcut is to add a hinge loss over random pairs of examples to each client's objective, penalising pairs where the feature increases but the prediction falls, and to measure success by the fraction of such ``violating'' random pairs (a random-pair monotonic violation rate, MVR). Monotonicity hints \citep{sill1996hints} avoid the problem by using pairs that differ \emph{only} in the constrained feature. We are not aware of published work that relies on the random-pair variant, and we include it as a cautionary baseline because it is easy to adopt by accident.
We show that this metric does not measure monotonicity, and that the loss can make violations worse. The basic question of how monotone federated models are, and what actually enforces monotonicity in FL, needs a different methodology.

\paragraph{Key observation.} Monotonicity is a property of the \emph{function}, not of the data. Whether $s_m\,\partial f/\partial x_m \ge 0$ holds at a point $x$ can be checked, penalised and certified at any point of the input domain, including points no client owns. If the input domain $\X$ is known to all parties (we treat it as an assumption: a public schema of feature ranges and categories, or bounds released with differential privacy or secure multi-party computation), then the server can audit the global model, and clients can penalise it on points beyond their own data, without exchanging any data.

\paragraph{Contributions.}
\begin{enumerate}
\item \textbf{A measurement caution.} The random-pair MVR compares points that differ in \emph{all} features, so it measures marginal rank agreement rather than monotonicity. A strictly monotone function can score arbitrarily close to $1/2$ (\cref{prop:mvr}). In our runs a monotone-by-construction model scores the same as FedAvg, and the random-pair rate is at most weakly correlated with ceteris-paribus violations (\cref{sec:res-metric}). We use gradient-sign and finite-difference violation rates on the data distribution and over the input domain, with a Clopper--Pearson bound on the violated volume (\cref{prop:cp}).
\item \textbf{Monotonicity and aggregation.} Any aggregator that forms convex combinations of parameters, FedAvg included, preserves monotonicity whenever the monotone parameters form a convex set or the model is reparameterised to be monotone everywhere (\cref{prop:convex}). Monotone architectures are therefore directly usable in FL, whereas unconstrained networks enjoy no such closure (\cref{ex:average}). We also make precise what a penalty evaluated on client data does and does not control (\cref{prop:support}).
\item \textbf{Enforcement beyond the clients' data.} Gradient penalties that add samples from the input box to each client's minibatch, and \textbf{CEGIS-FL}, in which the server searches the global model for verified violations and broadcasts them as additional penalty points. The added points need no client data.
\item \textbf{A careful benchmark.} \nRunsMain{} runs on six tasks, with $\lambda$ selected on validation data only, 5--10 seeds, paired tests, matched-$\lambda$ dose--response curves and an independent re-evaluation. Enforcement does not cost AUC: equivalence to FedAvg within $\pm0.01$ is established in \nEquiv{} of \nEquivTot{} method--task comparisons, and the remaining one is inconclusive rather than worse. On the data distribution, penalties on client data and on box samples are comparable, with neither consistently lower. Over the domain, box samples leave fewer violations at every strength (point estimates of 2.5 to more than 10 times less $\lambda$ for the same level, with wide intervals on the small tasks). Ablations attribute this to the location of the points rather than their number. CEGIS-FL reaches low violation levels at 10--100 times smaller $\lambda$; on two tasks Grad-Box does not reach them within our grid. The random-pair hinge loss is either ineffective or costly in AUC. We also report what did not work: a server-side correction step and mixup points.
\end{enumerate}

The paper is an analysis-and-tools paper. Its ingredients exist in the centralized literature: pointwise gradient penalties \citep{gupta2019pointwise,monteiro2022regularization}, regularisation on uniform samples from the domain \citep{liu2020certified}, a comparison of where such penalties are evaluated \citep{monteiro2022regularization}, counterexample-guided training \citep{sivaraman2020comet} and constrained networks. We credit them in \cref{sec:related}. Our contribution is to bring them to FL, where the domain must be agreed on without pooling data and the server is the natural auditor. We show why random-pair evaluation is misleading, measure systematically which choices matter, and report effects that are \emph{not} specific to federation. Centralized FedAvg violates the rules at least as much as federated FedAvg on most tasks, the advantage of box points persists in IID partitions, and on pooled data it persists on three of six tasks. What federation changes is that box samples are the only additional penalty points available without data.

\section{Related work}
\label{sec:related}

\paragraph{Monotone models.}
\emph{Architectural} approaches guarantee monotonicity by construction. Examples are sign-constrained or min--max networks \citep{sill1997monotonic,daniels2010monotone}, calibrated lattices and deep lattice networks \citep{gupta2016lattice,you2017dln}, integral-based networks \citep{wehenkel2019umnn}, constrained networks that mix convex and concave activations \citep{runje2023cmnn}, Lipschitz-constrained networks \citep{nolte2023expressive} and scalable variants \citep{kim2024scalable}.
\emph{Regularisation} approaches add a penalty instead. Monotonicity hints \citep{sill1996hints} penalise pairs of virtual examples. Pointwise penalties on the partial derivative \citep{gupta2019pointwise,monteiro2022regularization} are the closest to ours.
\citet{liu2020certified} train with a penalty on random points and then certify with a MILP. \citet{sivaraman2020comet} use counterexamples found by an SMT solver.
\citet{monteiro2022regularization} also study \emph{where} the penalty is evaluated, comparing penalties on training points, random points and mixtures of the two. Our federated results on penalty location are in line with this centralized line of work.
All of these methods are centralized. Our gradient penalty and our counterexample loop adapt them. The federated questions are how the domain is agreed on without pooling data, which points each client can penalise, and what the server can audit on its own.

\paragraph{Knowledge injection and neuro-symbolic learning.}
Semantic losses \citep{xu2018semantic} and DL2 \citep{fischer2019dl2} turn logical constraints into differentiable penalties (see \citealp{garcez2023neurosymbolic} for a survey).
\citet{xing2024fednsl} study federated neuro-symbolic \emph{rule learning}, which is a different problem from enforcing known feature-level monotonicity.

\paragraph{Constrained and fair federated learning.}
Federated optimisation under heterogeneity \citep{li2020fedprox,karimireddy2020scaffold,hsu2019noniid} targets accuracy and convergence.
Constraint-aware FL mostly targets fairness \citep{mohri2019agnostic,cui2021fcfl,ezzeldin2023fairfed} or convex constraints with augmented-Lagrangian methods \citep{he2024convex}.
Monotonicity constraints differ: they are pointwise over the input domain and non-convex in the parameters of a neural network.
Concurrent work on federated additive models with sign-constrained spline increments \citep{fedadditive2026} observes, for additive models, that averaging preserves the monotone direction. This is a special case of our \cref{prop:convex}.

\paragraph{Verification.}
Certifying that a ReLU network is monotone is a verification problem over its derivative. We use MILP encodings \citep{tjeng2019milp,liu2020certified}, interval bounds \citep{gowal2018ibp} and branch-and-bound \citep{bunel2020bab}. Tighter linear relaxations \citep{zhang2018crown} would extend the reach of exact certification. Here we rely on statistical certificates for unconstrained networks and exact guarantees for constrained ones.

\section{Problem setup}
\label{sec:setup}

\paragraph{Federated learning.}
$K$ clients hold datasets $D_k$ drawn from distributions $P_k$ over $\X\times\{0,1\}$. FedAvg \citep{mcmahan2017fedavg} minimises $\sum_k p_k\,\mathbb{E}_{P_k}[\ell(f_\theta(x),y)]$ with $p_k \propto |D_k|$. In each round, the selected clients run a few local optimiser steps from the global parameters, and the server replaces them with the weighted average $\theta \leftarrow \sum_k p_k \theta_k$. We write $P=\sum_k p_k P_k$ for the pooled input distribution, which no single party can access.

\paragraph{Assumption: a shared input domain.}
Inputs are numeric features with bounds $[l_j,u_j]$ and categorical features with finite vocabularies, so $\X=\prod_j [l_j,u_j]\times\prod_g \mathcal{C}_g$. We assume this domain is known to all parties. Federated preprocessing needs it anyway, together with moments for standardisation.
Counts, sums, sums of squares and category counts are sums over clients, so secure aggregation can compute them without revealing any client. \emph{Minima and maxima are not sums.} They have to come from a public schema (e.g.\ admissible ranges in a data dictionary), from secure multi-party computation, or from a differentially private release. Exact extremes can also be identifying (a single maximum capital gain is a property of one person), so we recommend schema bounds or DP-noised bounds in practice (\cref{sec:discussion}).
In our experiments the bounds are the exact pooled training minima and maxima, computed from per-client statistics; the pooled transform is reproduced exactly (a unit test checks this). The resulting box $\X$ and a sampler $\U(\X)$ (uniform over the numeric box, with categories drawn from their aggregate frequencies) are available to every party.

\begin{definition}[Partial monotonicity]
\label{def:mono}
A rule $(m,s_m)$, with $s_m\in\{+1,-1\}$, holds for $f:\X\to\mathbb{R}$ (the logit) if $s_m\,(f(x+\delta e_m)-f(x))\ge 0$ for all $x\in\X$ and $\delta>0$ with $x+\delta e_m\in\X$. For $f$ differentiable almost everywhere, as ReLU networks are, this is equivalent to $s_m\,\partial_m f(x)\ge 0$ for almost every $x\in\X$.
\end{definition}

\paragraph{Measuring violations ceteris paribus.}
Monotonicity concerns how $f$ changes when one feature moves and \emph{all others are held fixed}. For a set of rules $\mathcal{M}$, a tolerance $\tau$ (we use $\tau=10^{-4}$ on the logit scale) and a measure $\mu$ on $\X$, we define the \emph{violated fraction}
\[
\Vol_\mu(f)=\mu\{x:\ \exists m\in\mathcal{M},\ s_m\,\partial_m f(x)<-\tau\}.
\]
We report it under three measures that answer different questions.
$\Vol_P$, estimated on test points, asks ``how often is the rule broken where data lives?''.
$\Vol_\U$, estimated on uniform box samples, asks ``on what fraction of the admissible domain is the rule broken?''. It includes inputs that are rare or unseen in every client's data, to which a deployed model can still be applied. Because the box spans exact extremes, much of its volume can be far from the data (e.g.\ large capital gains), so $\Vol_\U$ weights extrapolation heavily.
We therefore also report $\Vol_{\U_q}$, on a \emph{data-supported} box whose numeric ranges are clipped to the 1st--99th percentiles of the test inputs.
None of these is ``the'' right measure. Which one matters depends on where the model will be queried. A finite-difference variant (a step of 5\% of the feature range) agrees closely with the gradient version in all our runs (\cref{app:additional}). Finally, \emph{certified} and \emph{violated volumes} are fractions of $\X$ on which monotonicity is \emph{proven} to hold or to fail (\cref{sec:verification}).

\paragraph{The random-pair MVR is not a monotonicity measure.}
A natural alternative samples random pairs $(x,x')$ from the data, keeps the pairs with $x_m>x'_m$, and counts how often $s_m(f(x)-f(x'))<0$. Because $x$ and $x'$ differ in every feature, this quantity measures how well the prediction agrees with feature $m$ in rank, \emph{marginally}, not monotonicity.

\begin{proposition}[Random-pair MVR]
\label{prop:mvr}
Let $X_m,X_j$ be independent with continuous distributions, and let $f(x)=a x_m + b x_j$ with $a>0$, $b\ge 0$. Then $f$ is strictly increasing in $x_m$. Yet its random-pair MVR, $\Pr[f(X)<f(X')\mid X_m>X'_m]$ for an independent copy $X'$, tends to $1/2$ as $b/a\to\infty$, and it is strictly positive whenever $b>0$ and $X_j$ is non-degenerate.
\end{proposition}

The proof is in \cref{app:proofs}. In this example the MVR is zero only if $b=0$, i.e.\ only if the prediction ignores every other feature. Penalising random-pair violations, which is what the pairwise hinge loss does, pushes the model to suppress other features or to bend the partial derivative of feature $m$. We observe both effects in \cref{sec:results}.

\section{Monotonicity under federated aggregation}
\label{sec:theory}

This section makes three points: when aggregation preserves monotonicity, what client-side penalties can and cannot control, and how a server can bound the violated volume without data. All proofs are in \cref{app:proofs}.

\begin{proposition}[Convex constraint sets survive aggregation]
\label{prop:convex}
Let $\Theta_{\mathcal M}\subseteq\Theta$ be convex and suppose $f_\theta$ satisfies all rules in $\mathcal M$ for every $\theta\in\Theta_{\mathcal M}$. If every client returns $\theta_k\in\Theta_{\mathcal M}$, then every convex combination $\sum_k p_k\theta_k$ lies in $\Theta_{\mathcal M}$. In particular, FedAvg and FedProx produce a monotone global model in every round. If instead $f_\theta$ is monotone for \emph{all} $\theta\in\Theta$ (a reparameterisation), no condition on the clients or on the server optimiser is needed.
\end{proposition}

The proposition is elementary, and we state it because it settles a practical question. It covers lattice models, whose monotonicity constraints are linear inequalities in the vertex parameters \citep{gupta2016lattice}, networks with non-negativity constraints on the paths from monotone inputs \citep{sill1997monotonic,daniels2010monotone,runje2023cmnn}, and the reparameterised network we use (\cref{sec:methods}). Additive models with sign-constrained increments \citep{fedadditive2026} are another special case.
For a constrained (not reparameterised) parameterisation, server optimisers with momentum or adaptive steps (FedAvgM, FedAdam) do \emph{not} form convex combinations and need a projection after the server step.
The cost of the guarantee is architectural, not federated.
Unconstrained networks enjoy no such closure:

\begin{example}[Averaging two monotone networks]
\label{ex:average}
Take one-unit networks $f(x)=v\,\relu(wx+b)$. With $(w,b,v)=(2,0,1)$, $f_A(x)=\relu(2x)$ is non-decreasing. With $(w,b,v)=(-1,2,-3)$, $f_B(x)=-3\,\relu(2-x)$ is non-decreasing. Their parameter average $(\tfrac12,1,-1)$ gives $f(x)=-\relu(x/2+1)$, which is strictly \emph{decreasing} for $x>-2$.
\end{example}

So, for unconstrained networks, monotonicity has to be re-established on the aggregate, and the question is which points the constraint is enforced on.
Let $\rho(f,x)=\sum_{m\in\mathcal M}\relu\!\big(-s_m\,\partial_m f(x)\big)^2$ be the pointwise penalty and $R_\mu(f)=\mathbb{E}_{x\sim\mu}\,\rho(f,x)$ its average under a measure $\mu$.

The next statement is not a quantitative prediction. It makes precise an intuition about what a penalty can and cannot see.

\begin{proposition}[What a penalty controls]
\label{prop:support}
\textbf{(a)} If client $k$ penalises its own data, the federated objective penalises $\sum_k p_k R_{P_k}=R_P$: the pooled \emph{data} distribution, not the domain.
\textbf{(b)} For any $\mu$ and $\tau>0$, the violated fraction at margin $\tau$ satisfies $\Vol^\tau_\mu(f)\le R_\mu(f)/\tau^2$.
\textbf{(c)} If $\U\ll P$ with density ratio bounded by $r=\|d\U/dP\|_\infty$, then $\Vol_\U(f)\le r\,\Vol_P(f)$. Without this condition there is no such bound: $R_P(f)$ depends on $f$ only through its derivatives on $\operatorname{supp}(P)$, so functions that agree there but differ elsewhere have the same data penalty, while their $\Vol_\U$ can differ by up to $\U(\X\setminus\operatorname{supp}P)$.
\end{proposition}

Caveats: (a) describes the federated \emph{objective}, which FedAvg with local steps only approximately minimises under client drift; the Markov bound in (b) is loose for small $\tau$; the density ratio in (c) is unbounded wherever the box contains regions without data; and the second statement is vacuous if $\operatorname{supp}P=\X$. Its practically relevant form is finite-sample: each client penalises $f$ only at its own training points and leaves it free in between, and the constructed functions need not be reachable by training.
The proposition therefore explains \emph{why} a penalty should be evaluated where violations are measured, not by how much. A penalty on $P$ targets violations under $P$; a penalty on $\U$ targets violations under $\U$. Data are sparse in the tails of numeric features and in rare category combinations, where $r$ is large, so we expect data penalties to need a larger $\lambda$ for the same effect on $\Vol_\U$, but not on $\Vol_P$. \Cref{sec:res-efficiency} confirms both halves of this expectation empirically. The size of the gap (2.5 to more than 10 times) is a measurement, not a consequence of the proposition.

\begin{proposition}[Statistical certificate from box samples]
\label{prop:cp}
Let $x_1,\dots,x_N\sim\U(\X)$ be i.i.d. and independent of $f$, and let $k$ be the number of samples on which some rule is violated (each check is an exact evaluation of $\partial_m f(x_i)$). Then, with probability at least $1-\delta$, $\Vol_\U(f)\le B^{-1}(1-\delta;\,k+1,\,N-k)$, the one-sided Clopper--Pearson bound \citep{clopper1934}. For $k=0$ this is $1-\delta^{1/N}\approx\ln(1/\delta)/N$, i.e.\ $0.15\%$ for $N=2000$ and $\delta=0.05$.
\end{proposition}

A server can compute this certificate for the global model with no client data. The bound holds for each model separately at level $1-\delta$; it is not simultaneous over the many models we report, which would need a union bound. It concerns violations at margin $\tau$ under $\U$, and says nothing about the worst case. It complements exact certificates, which are available by construction for constrained networks and, for small unconstrained ReLU networks, from branch-and-bound or MILP verification (\cref{sec:verification}).

\section{Methods}
\label{sec:methods}

All methods share the FedAvg protocol and differ only in the client objective, the architecture and, for CEGIS-FL, a server-side search for counterexamples.

\paragraph{Monotone by construction (MonoNet).}
We reparameterise a ReLU-style MLP so that every parameter value gives a monotone function. In the first layer, the columns acting on a monotone feature $m$ are replaced by $s_m|W_{:,m}|$. All later weight matrices are replaced by $|W|$. Following \citet{runje2023cmnn}, each hidden layer applies, to equal thirds of its units, a convex ($\relu(z)$), a concave ($-\relu(-z)$) and a saturating ($\operatorname{hardtanh}(z)$) monotone activation, so that a network with non-negative weights is not restricted to convex functions of the monotone inputs. Unconstrained features enter the first layer freely. By \cref{prop:convex}, the global model is monotone after every FedAvg round whatever the clients do.

\paragraph{Pointwise penalty and where it is evaluated.}
For an unconstrained MLP $f_\theta$, client $k$ minimises
\begin{equation}
\label{eq:client}
\frac{1}{|B|}\sum_{(x,y)\in B}\ell\big(f_\theta(x),y\big)\;+\;\lambda\,\frac{1}{|Z|}\sum_{z\in Z}\rho(f_\theta,z),
\qquad \rho(f,z)=\sum_{m\in\mathcal M}\relu\!\big(-s_m\,\partial_m f(z)\big)^2,
\end{equation}
where $B$ is the minibatch and $Z$ is the set of \emph{penalty points}. The gradient $\partial_m f(z)$ is computed by automatic differentiation; minimising the penalty requires a second backward pass. The methods differ only in $Z$:
\begin{itemize}
\item \textbf{Grad-Data}: $Z=B$, the client's own inputs (the federated version of \citealp{gupta2019pointwise,monteiro2022regularization}).
\item \textbf{Grad-Mixup}: $B$ plus random convex combinations of pairs in $B$, which cover the convex hull of the client's data.
\item \textbf{Grad-Box}: $B$ plus 64 fresh samples from $\U(\X)$ per step. The added points are \emph{data-free}: every client can draw them from the shared box.
\item \textbf{CEGIS-FL}: Grad-Box plus up to 64 points sampled from a broadcast counterexample pool $C$ per step (\cref{alg:cegis}).
\end{itemize}
The methods therefore differ in the \emph{number} of penalty points as well as in their location: with minibatch size $|B|=64$ (32 on Heart), Grad-Box penalises 2$\times$ (3$\times$ on Heart) as many points per step as Grad-Data, and CEGIS-FL up to 3$\times$ (5$\times$). Because \cref{eq:client} averages $\rho$ over $Z$, adding points mainly reweights the penalty between locations and reduces its variance, rather than scaling it. To separate location from count, we add two ablations: \textbf{Box-only}, with $Z$ = 64 box samples and no client points, and \textbf{Grad-Data$\times$2}, with $Z=B$ plus 64 further client examples per step, which matches Grad-Box's point count with client data (\cref{sec:res-ablation}).

\begin{algorithm}[t]
\caption{CEGIS-FL (server side; clients minimise \cref{eq:client} with $Z=B\cup Z_{\U}\cup Z_C$)}
\label{alg:cegis}
\begin{algorithmic}[1]
\State $C\gets\emptyset$
\For{round $r=1,\dots,R$}
  \State broadcast $\theta$ and $C$; clients train locally with penalty points $B$, $Z_\U\sim\U(\X)$, $Z_C\sim C$
  \State $\theta\gets\sum_k p_k\theta_k$ \Comment{FedAvg}
  \If{$r \bmod T = 0$}
    \For{each rule $(m,s_m)$}
      \State draw $N$ points from $\U(\X)$; keep the $4c$ with the most negative $s_m\partial_m f_\theta$
      \State refine them by random local search inside $\X$ (20 steps, shrinking radius)
      \State keep up to $c$ points with $s_m\partial_m f_\theta<-\tau$ (checked by autograd)
    \EndFor
    \State add the kept points and $J$ jittered copies of each (numeric features only) to $C$; cap $|C|$
  \EndIf
\EndFor
\end{algorithmic}
\end{algorithm}

The counterexamples are functions of the global model and the public box only, so broadcasting them reveals nothing about client data. CEGIS-FL adapts counterexample-guided learning \citep{sivaraman2020comet} to FL. Its search is a cheap falsifier rather than a solver, because the counterexamples only need to be \emph{verified} violations, not the worst ones.
We also evaluate \textbf{Grad-Box + server}: after each aggregation the server takes a few gradient steps on $\lambda_s R_\U(f_\theta)+\tfrac{\mu_s}{2}\|\theta-\theta_{\text{agg}}\|^2$. This uses no data but also no task loss.

\paragraph{Pairwise baseline.}
\textbf{Pairwise} adds $\lambda\cdot\mathrm{mean}\,\relu\!\big(s_m(f(x_j)-f(x_i))\big)$ over random pairs from the minibatch with $x_{i,m}>x_{j,m}$, with all other features free. This is the loss counterpart of the random-pair MVR.

\subsection{Server-side verification}
\label{sec:verification}
Beyond the statistical certificate of \cref{prop:cp}, we implement two sound verifiers for ReLU MLPs. Neither needs data.
\emph{(i)} An exact MILP. It encodes each unstable ReLU with big-M constraints, the products $\delta\cdot g$ in the derivative chain $\partial_m f=W_LD_{L-1}\cdots D_1W_{1,:m}$ with McCormick constraints, and categorical features as exact one-hot groups. All constants are tightened by LP-based bound tightening. Following \citet{liu2020certified}, but for any depth, it decides whether $s_m\partial_m f\le-\tau$ is feasible anywhere on $\X$, and any counterexample is moved into the interior of its linear region so that autograd reproduces it exactly.
\emph{(ii)} Input-domain branch-and-bound with interval bounds on the activations and on the derivative chain \citep{gowal2018ibp,bunel2020bab}. On a sub-box where all neurons are stable the derivative is constant, so the bound is exact. The output is a verdict and the \emph{proven} certified and violated volume fractions.
Both are cross-checked against dense sampling in our test suite (\cref{app:details}). For unconstrained networks of the size used here (two hidden layers of 32--64 units), almost every neuron is unstable over the whole box. In preliminary runs, the MILP did not terminate within 90\,s per rule even for 80 binary variables, so the results in \cref{sec:res-cert} use the falsifier and branch-and-bound only. Longer budgets, tighter relaxations \citep{zhang2018crown} or training aimed at certifiability \citep{liu2020certified} may well certify penalised networks; we did not pursue this. We therefore use \cref{prop:cp} as the main certificate for unconstrained models.

\section{Experimental setup}
\label{sec:experiments}

\begin{table}[t]
\caption{Tasks, federations and monotonicity rules. Signs refer to the logit of $P(y=1)$. Each rule's rationale is in \cref{app:datasets}. ``Natural'' splits use the real data owners as clients.}
\label{tab:datasets}
\centering
\small
\resizebox{\linewidth}{!}{%
\begin{tabular}{lrllll}
\toprule
Task & $n$ & Clients & Partition & Rules & Rounds \\
\midrule
Synthetic & 20{,}000 & 10 & covariate skew on $x_1$ & $x_1\!+$, $x_2\!+$ (ground truth) & 100 \\
Adult & 48{,}842 & 10 & Dirichlet label skew, $\alpha=0.3$ & education, hours/week, capital gain $(+)$ & 40 \\
COMPAS & 6{,}172 & 5 & Dirichlet label skew, $\alpha=0.3$ & priors, three juvenile counts $(+)$ & 100 \\
German & 1{,}000 & 3 & Dirichlet label skew, $\alpha=1$ & duration, credit amount $(-)$ & 100 \\
Heart & 920 & 4 & natural (4 hospitals) & age, rest.\ BP, ST depr.\ $(+)$; max HR $(-)$ & 80 \\
ACS Income & 69{,}833 & 10 & natural (10 US states) & education, hours/week $(+)$ & 100 \\
\bottomrule
\end{tabular}}
\end{table}

\paragraph{Tasks.}
\Cref{tab:datasets} summarises the six tasks.
\emph{Synthetic} has a known logit that is increasing in $x_1,x_2$, with deliberately \emph{weak} monotone effects relative to non-monotone terms in $x_3$--$x_5$ and a categorical feature (\cref{app:datasets}). Clients own disjoint quantile bands of $x_1$.
\emph{Adult} \citep{kohavi1996adult}, \emph{COMPAS} \citep{angwin2016machine} (standard filtering; race excluded) and \emph{German credit} \citep{hofmann1994german} are partitioned with Dirichlet label skew \citep{hsu2019noniid}.
\emph{Heart} \citep{detrano1989heart} uses its four contributing hospitals as clients, the split used by FLamby \citep{terrail2022flamby}.
\emph{ACS Income} \citep{ding2021retiring} uses ten US states as clients: seven states with more than 8{,}000 people are subsampled to 8{,}000, and three smaller states are kept whole (\cref{app:datasets}).
We do not constrain age on Adult, because income is non-monotone in age; a rule there would be misspecified.

\paragraph{Models and training.}
Two-hidden-layer MLPs (64 units; 32 on German and Heart) and MonoNet of the same width. Adam with learning rate $10^{-3}$, batch size 64 (32 on Heart), one local epoch per round, all clients every round.
Preprocessing is computed from client sufficient statistics (\cref{sec:setup}). Each client splits its data into 70/10/20 train/validation/test. Validation and test sets are the unions of the clients' splits.
Round budgets come from a calibration study (\cref{app:protocol}): validation AUC peaks early on Adult and late on COMPAS and ACS.
Because violations keep growing after AUC saturates (\cref{sec:res-baseline}), we evaluate every 5 rounds (every 2 on Adult) and keep the checkpoint with the best \emph{validation AUC}. Last-round results are reported in \cref{app:additional}.

\paragraph{Selection protocol.}
Grad-Data and Grad-Box are swept over $\lambda\in\{0.1,1,10,30,100,300,1000\}$ and CEGIS-FL over $\{0.1,1,10,30,100\}$. Pairwise is swept over $\{0.01,0.1,1,10\}$, because larger values already cost substantial AUC. Grad-Mixup and Grad-Box + server are swept over $\{0.1,1,10\}$. For each method and task we select the $\lambda$ with the lowest validation violated fraction among those whose mean validation AUC is within 0.005 of FedAvg's. The violated fraction used for selection is the in-run $\Vol_\U$ on a fixed set of 2{,}000 box samples, drawn from a random stream separate from the box samples used in training. Test data never enter selection.
Validation AUC stays flat up to $\lambda=1000$ (\cref{sec:res-acc}), so the AUC constraint rarely binds. The selected $\lambda$ for Grad-Data and Grad-Box is 300 or 1000 in 11 of 12 cases, and 8 of these are at the top of the grid (\cref{tab:lambda}). CEGIS-FL selects below the top of its grid on three tasks. Selected-$\lambda$ comparisons therefore partly reflect the grid, and Grad-Box and CEGIS-FL are not compared at equal $\lambda$ there. Our main evidence comparing penalty locations is the \emph{matched-$\lambda$} analysis (\cref{sec:res-efficiency}), which involves no selection.
\emph{Ablations} (\cref{sec:res-ablation}) at $\lambda\in\{10,100\}$ separate the location of the penalty points from their number, and repeat the comparison on pooled data.
We run 5 seeds (10 for German and Heart) and pair methods by seed, which fixes the partition and initialisation. \Cref{tab:runs} breaks down the \nRunsMain{} runs.
Penalty-free baselines are FedAvg and MonoNet. FedAvg and MonoNet trained on the pooled data (``centralized'') are reference points.

\paragraph{Metrics.}
AUC is the primary accuracy metric. Threshold metrics at 0.5 are unstable under label skew for \emph{every} method; for example, the predicted-positive rate on Adult is 0.12 against a base rate of 0.24. We report balanced accuracy and F1 in \cref{app:additional}.
AUC parity is partly built into the selection rule (validation AUC within 0.005 of FedAvg). We therefore also report AUC at matched $\lambda$, where no selection is involved, and paired equivalence tests (TOST at $\alpha=0.05$, margin $\pm0.01$, equivalent to a 90\% confidence interval inside the margin). We also check the stricter criterion that the 95\% interval lies inside the margin.
Violations are measured as in \cref{sec:setup} with $\tau=10^{-4}$: $\Vol_P$ on test points, $\Vol_\U$ on uniform box samples, and $\Vol_{\U_q}$ on the data-supported box.
Final $\Vol_\U$ and $\Vol_{\U_q}$ values and certificates use 10{,}000 \emph{fresh} box samples per model, drawn with a seed never used in training or selection, so that \cref{prop:cp} applies.
We also report the random-pair MVR on test data, to compare it with the ceteris-paribus metrics.
All experiments ran on a CPU-only workstation. Run times are reported in \cref{app:additional}.

\section{Results}
\label{sec:results}

\begin{table}[t]
\caption{Violated fraction of the input domain $\Vol_\U$ (\%, mean $\pm$ s.d.\ over seeds; lower is better) at the validation-selected $\lambda$, measured on 10{,}000 fresh box samples per model that were not used in training or selection. Bold: lowest among penalty and penalty-free methods (MonoNet is 0 by construction). Bottom rows: training on pooled data.}
\label{tab:violation}
\centering
\small
\begin{tabular}{lcccccc}
\toprule
Method & Synthetic & Adult & COMPAS & German & Heart & ACS Income \\
\midrule
FedAvg & 11.8{\scriptsize$\pm$2.5} & 26.1{\scriptsize$\pm$20.9} & 86.9{\scriptsize$\pm$16.4} & 7.3{\scriptsize$\pm$7.7} & 50.1{\scriptsize$\pm$27.9} & 26.1{\scriptsize$\pm$4.9} \\
Pairwise & 10.1{\scriptsize$\pm$2.5} & 4.8{\scriptsize$\pm$2.8} & 57.3{\scriptsize$\pm$37.7} & 8.0{\scriptsize$\pm$9.8} & 32.6{\scriptsize$\pm$20.4} & 26.0{\scriptsize$\pm$4.7} \\
Grad-Data & 0.1{\scriptsize$\pm$0.0} & 0.0{\scriptsize$\pm$0.1} & 0.5{\scriptsize$\pm$0.8} & 1.3{\scriptsize$\pm$2.0} & 4.8{\scriptsize$\pm$4.0} & 0.5{\scriptsize$\pm$0.3} \\
Grad-Box & 0.0{\scriptsize$\pm$0.0} & \textbf{0.0}{\scriptsize$\pm$0.0} & \textbf{0.0}{\scriptsize$\pm$0.0} & 0.3{\scriptsize$\pm$0.3} & 0.6{\scriptsize$\pm$0.2} & 0.0{\scriptsize$\pm$0.0} \\
CEGIS-FL & \textbf{0.0}{\scriptsize$\pm$0.0} & 0.0{\scriptsize$\pm$0.0} & 0.0{\scriptsize$\pm$0.0} & \textbf{0.0}{\scriptsize$\pm$0.0} & \textbf{0.1}{\scriptsize$\pm$0.0} & \textbf{0.0}{\scriptsize$\pm$0.0} \\
\midrule
MonoNet & 0.0{\scriptsize$\pm$0.0} & 0.0{\scriptsize$\pm$0.0} & 0.0{\scriptsize$\pm$0.0} & 0.0{\scriptsize$\pm$0.0} & 0.0{\scriptsize$\pm$0.0} & 0.0{\scriptsize$\pm$0.0} \\
\midrule
FedAvg (centralized) & 22.0{\scriptsize$\pm$12.4} & 48.6{\scriptsize$\pm$27.5} & 83.1{\scriptsize$\pm$19.3} & 7.7{\scriptsize$\pm$9.2} & 64.1{\scriptsize$\pm$24.9} & 26.6{\scriptsize$\pm$7.1} \\
MonoNet (centralized) & 0.0{\scriptsize$\pm$0.0} & 0.0{\scriptsize$\pm$0.0} & 0.0{\scriptsize$\pm$0.0} & 0.0{\scriptsize$\pm$0.0} & 0.0{\scriptsize$\pm$0.0} & 0.0{\scriptsize$\pm$0.0} \\
\bottomrule
\end{tabular}

\end{table}

\subsection{Unconstrained federated models break domain rules on much of the domain}
\label{sec:res-baseline}
FedAvg violates at least one rule on \vFedAvgcompas\% of the input box for COMPAS, \vFedAvgheart\% for Heart, \vFedAvgacsincome\% for ACS Income, \vFedAvgadult\% for Adult, \vFedAvgsynthetic\% for Synthetic and \vFedAvggerman\% for German (\cref{tab:violation}).
Counting only clearly wrong-signed slopes of at least $0.1$ logit per standard deviation of the feature, FedAvg still violates the rules on 46\% of the box for COMPAS, 27\% for Heart, 10\% for ACS Income, 5\% for Synthetic and 4\% for Adult (0.4\% for German), while all gradient penalties are at or near 0 (\cref{tab:magnitude}).
Violations are not a transient of early training. On every task, after the first few rounds, FedAvg's violated fraction increases with further training (\cref{fig:convergence-all}). On Adult it rises from 19\% after two rounds to 34\% after 40, while test AUC peaks after six rounds and then declines slowly (\cref{fig:convergence}).
Pooled training does \emph{not} remove the problem: centralized FedAvg violates more on Adult, Heart and Synthetic (e.g.\ 49\% vs.\ 26\% on Adult), about as much on ACS Income and German, and slightly less on COMPAS. We therefore do \emph{not} claim that federation causes violations; it removes the usual remedy of inspecting and constraining the model with pooled data.

\subsection{Enforcement costs no measurable accuracy}
\label{sec:res-acc}
Every gradient-penalty method and MonoNet is within $\pm0.005$ AUC of FedAvg on every task; the largest difference, $+0.005$ (Grad-Box on German), is an improvement (\cref{tab:auc}). Non-significance alone would not show parity, so we test equivalence directly. Equivalence to FedAvg within $\pm0.01$ AUC holds by TOST ($\alpha=0.05$) in \nEquiv{} of \nEquivTot{} method--task comparisons, and in the same 29 the stricter 95\% interval also lies inside the margin (\cref{tab:auc-ci}). The exception is MonoNet on German (interval $[-0.014,+0.012]$, 1{,}000 examples), which is inconclusive rather than worse. No difference is significant after Bonferroni correction.
Because the selection rule itself requires validation AUC within 0.005 of FedAvg, parity at the selected $\lambda$ is partly by design. The stronger evidence is at \emph{matched} $\lambda$, where no selection is involved: validation AUC is flat from $\lambda=0.1$ to $1000$ for Grad-Data and Grad-Box and to $100$ for CEGIS-FL on every task; across $\lambda$ the mean validation AUC of any method on any task varies by at most $0.0045$ (German, CEGIS-FL, where it \emph{increases} with $\lambda$).
The only accuracy cost we observe comes from the pairwise loss: $0.008$ AUC on Synthetic already at $\lambda=0.1$ ($p<0.001$), $0.008$--$0.24$ at $\lambda=1$ and $0.024$--$0.26$ at $\lambda=10$ (\cref{app:additional}), hence its selected $\lambda\le0.1$. Threshold metrics are less uniform (e.g.\ MonoNet's F1 on COMPAS; \cref{app:additional}).

\subsection{Where the penalty is evaluated: on the data vs.\ over the domain}
\label{sec:res-efficiency}

\begin{table}[t]
\caption{Violations where data lives: violated fraction on \emph{test points} $\Vol_P$ (\%) at the selected $\lambda$, mean $\pm$ s.d.\ over seeds. The data-supported box is in \cref{tab:clip}.}
\label{tab:vp}
\centering\scriptsize
\begin{tabular}{lcccccc}
\toprule
 & Synthetic & Adult & COMPAS & German & Heart & ACS Income \\
\midrule
FedAvg & 9.93{\scriptsize$\pm$2.83} & 5.37{\scriptsize$\pm$2.74} & 61.47{\scriptsize$\pm$39.66} & 1.99{\scriptsize$\pm$2.66} & 49.35{\scriptsize$\pm$31.09} & 4.87{\scriptsize$\pm$0.70} \\
Pairwise & 8.51{\scriptsize$\pm$2.95} & 3.26{\scriptsize$\pm$1.95} & 35.46{\scriptsize$\pm$16.70} & 2.14{\scriptsize$\pm$2.18} & 26.70{\scriptsize$\pm$19.07} & 3.82{\scriptsize$\pm$0.40} \\
Grad-Data & 0.05{\scriptsize$\pm$0.04} & 0.00{\scriptsize$\pm$0.00} & 0.92{\scriptsize$\pm$0.62} & 0.35{\scriptsize$\pm$0.78} & 1.41{\scriptsize$\pm$0.77} & 0.01{\scriptsize$\pm$0.01} \\
Grad-Box & 0.05{\scriptsize$\pm$0.04} & 0.03{\scriptsize$\pm$0.04} & 0.58{\scriptsize$\pm$0.32} & 0.10{\scriptsize$\pm$0.31} & 0.76{\scriptsize$\pm$0.58} & 0.01{\scriptsize$\pm$0.01} \\
CEGIS-FL & 0.07{\scriptsize$\pm$0.08} & 0.33{\scriptsize$\pm$0.62} & 1.33{\scriptsize$\pm$1.13} & 0.05{\scriptsize$\pm$0.16} & 0.16{\scriptsize$\pm$0.36} & 0.01{\scriptsize$\pm$0.01} \\
MonoNet & 0.00{\scriptsize$\pm$0.00} & 0.00{\scriptsize$\pm$0.00} & 0.00{\scriptsize$\pm$0.00} & 0.00{\scriptsize$\pm$0.00} & 0.00{\scriptsize$\pm$0.00} & 0.00{\scriptsize$\pm$0.00} \\
\bottomrule
\end{tabular}

\end{table}

\begin{figure}[t]
\centering
\includegraphics[width=\linewidth]{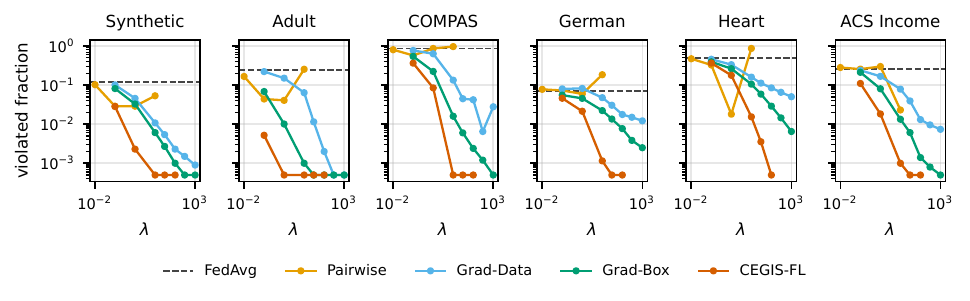}
\caption{Dose--response at matched penalty strength: violated fraction of the input box against $\lambda$ (log--log, mean over seeds, floored at $5\cdot10^{-4}$ for display; in-run evaluation on 2{,}000 box samples, with no selection involved). At every $\lambda$ on every task, CEGIS-FL $\le$ Grad-Box $<$ Grad-Data (CEGIS-FL up to $\lambda=100$), with equal validation AUC. FedAvg: dashed.}
\label{fig:lambda}
\end{figure}

\paragraph{On the data, neither location is consistently better.}
On test points at the selected $\lambda$ (\cref{tab:vp}), Grad-Box is lower on three tasks (COMPAS 0.58\% vs.\ 0.92\%, German 0.10\% vs.\ 0.35\%, Heart 0.76\% vs.\ 1.41\%), Grad-Data on Adult (0.00\% vs.\ 0.03\%), and the two are equal on Synthetic and ACS Income. At matched $\lambda$ and matched point count, the seed-level comparison on test points is close to even (\cref{sec:res-ablation}). We did not test equivalence, so we call them comparable rather than equal. Both reduce FedAvg's on-distribution violations by a factor of at least 5.

\paragraph{Over the domain, box points are more efficient.}
The location matters for $\Vol_\U$. Because selection is limited by the grid (\cref{sec:experiments}), we compare at \emph{equal} $\lambda$, which involves no selection (\cref{fig:lambda}).
In all \nCells{} task--$\lambda$ combinations up to $\lambda=100$ the mean $\Vol_\U$ of Grad-Box is below that of Grad-Data (\nBoxBeatsData/\nCells), and CEGIS-FL is at or below Grad-Box (\nCegisLeqBox/\nCells, strictly in 29), at equal validation AUC. Grad-Box also stays below Grad-Data at $\lambda=300$ and $1000$ (42 of 42 cells in total).
At the seed level, Grad-Box is lower in 168 of 200 paired comparisons up to $\lambda=100$ and in 62 of 80 at $\lambda\in\{300,1000\}$.
To quantify the gap, we interpolate each task's Grad-Data curve (log--log) to find the $\lambda$ that matches Grad-Box's violation at $\lambda=10$, with a seed bootstrap (\cref{tab:shift}). Grad-Data needs 2.5$\times$ (Synthetic; 95\% CI 1.1--5.1), 3.9$\times$ (Heart; 0.9--17), 6.0$\times$ (German; 0.6 to $>$100), 9.8$\times$ (ACS Income; 5.6--24), 15$\times$ (Adult; 2.4 to $>$100) and more than 100$\times$ (COMPAS; lower bound 1.5) the penalty weight. On German and Heart the intervals include parity, so the efficiency gain is established on four of the six tasks.
Grad-Box does not, however, reach CEGIS-FL's level within the grid on every task: at $\lambda=1000$ it matches CEGIS-FL at $\lambda=100$ on Synthetic, Adult, COMPAS and ACS Income (all $\le$0.02\%), but remains 8--19 times higher on German (0.25\% vs.\ 0.03\%) and Heart (0.65\% vs.\ 0.04\%).
This matches the intuition of \cref{prop:support}: each penalty controls violations under the distribution it is evaluated on.

\paragraph{On the data-supported box, the gap narrows.}
Much of the full box lies far from any data. On the data-supported box (\cref{tab:clip}), FedAvg still violates the rules on 4--85\% of the volume. Grad-Box is below Grad-Data on COMPAS (2.1\% vs.\ 5.9\%), German, Heart and ACS Income, and within 0.01 percentage points of it on Synthetic and Adult, where both are below 0.1\%.
CEGIS-FL is lowest or tied on the data-supported box on five of six tasks (on Adult all three are $\le$0.04\%), and lowest on test points on German and Heart. On Synthetic, Adult and COMPAS its test-point violations exceed Grad-Box's (e.g.\ 1.3\% vs.\ 0.6\% on COMPAS); there it was selected at a $\lambda$ 30--100 times smaller than Grad-Box's (\cref{tab:lambda}). Seed-paired tests are in \cref{app:additional}; mixup points and our server-side step did not help.

\subsection{Location or count? Ablations}
\label{sec:res-ablation}
Grad-Box penalises more points per step than Grad-Data (\cref{sec:methods}). At $\lambda\in\{10,100\}$ we compare it with Grad-Data$\times$2, which adds 64 \emph{client} examples instead of 64 box samples, and with Box-only (\cref{tab:ablation}). Counts are seed pairs out of 40 (in-run evaluation).
\emph{Location, not count, drives the domain-level gain.} With equal point counts, Grad-Box has a lower $\Vol_\U$ than Grad-Data$\times$2 in \abBoxVsDataTwoUTen{} pairs at $\lambda=10$ and \abBoxVsDataTwoUHundred{} at $\lambda=100$. On test points the two are close (lower in \abBoxVsDataTwoPTen{} and \abBoxVsDataTwoPHundred{}). Doubling the client points does not help either: Grad-Data$\times$2 is lower than Grad-Data in only \abDataTwoVsDataUTen{} and \abDataTwoVsDataUHundred{} pairs on $\Vol_\U$.
\emph{Box samples alone are not enough.} Box-only is as good as Grad-Box over the domain (lower than Grad-Data$\times$2 in \abBoxOnlyVsDataTwoUTen{} and \abBoxOnlyVsDataTwoUHundred{} pairs), but worse on test points (lower in only \abBoxOnlyVsDataTwoPTen{} and \abBoxOnlyVsDataTwoPHundred{}). On COMPAS, 32\% of test points violate a rule at $\lambda=10$ and 22\% at $\lambda=100$, against 8\% and 2\% for Grad-Box. The client term controls violations where data lives; combining it with box samples is what makes Grad-Box good on both.
\emph{The effect is not specific to federation.} Trained on pooled data at $\lambda=100$, Grad-Box has a lower $\Vol_\U$ than Grad-Data in \cenBoxLtData{} seed pairs (\cref{tab:central}). The gap persists on COMPAS (0.75\% vs.\ 10.4\%), Heart (2.5\% vs.\ 10.9\%) and ACS Income (0.15\% vs.\ 2.4\%), and vanishes on Synthetic and Adult, where both are below 0.3\%. What federation changes is not the effect but its availability: box samples are the only additional penalty points that need no data.

\subsection{The random-pair loss and metric}
\label{sec:res-metric}

At the $\lambda$ it can afford in accuracy ($\le0.1$), the pairwise hinge loss stays close to FedAvg on ACS Income, German and Synthetic, and reduces violations only partially on Adult (4.8\% vs.\ 26.1\%), Heart (32.6\% vs.\ 50.1\%) and COMPAS (57.3\% vs.\ 86.9\%), far short of the gradient penalties (\cref{tab:violation}). At $\lambda=10$ it loses 0.02--0.26 AUC everywhere and \emph{increases} violations above FedAvg on Heart (87\% vs.\ 50\%), COMPAS (97\% vs.\ 87\%) and German (19\% vs.\ 7\%) (\cref{fig:lambda}). Both are consistent with the two ways of reaching marginal rank agreement in \cref{prop:mvr} (suppressing other features, bending the constrained derivative), though we did not isolate them.
The random-pair MVR does not track monotonicity (\cref{fig:mvr}). Within tasks, its Spearman correlation with $\Vol_\U$ is $-0.14$ on Adult, $0.01$ on COMPAS, $-0.07$ on ACS Income, $0.13$ on German, $0.36$ on Synthetic and $0.38$ on Heart. MonoNet has the same random-pair MVR as FedAvg (0.291 on Adult), and the pairwise runs that violate the rule on nearly the whole box obtain the \emph{lowest} (0.06--0.15).

\subsection{Heterogeneity}
\label{sec:res-het}
At a fixed $\lambda=10$ (3 seeds; \cref{fig:het}, \cref{app:additional}), covariate skew on a constrained feature raises FedAvg's violations on Synthetic (6.5\% IID vs.\ 12.3\%), whereas label skew on Adult has no consistent effect. Grad-Box stays at or below 0.6\% and CEGIS-FL and MonoNet at 0 in all eight settings; the advantage of box points is present at every level of heterogeneity, IID included.

\subsection{Certificates and cost}
\label{sec:res-cert}
\Cref{prop:cp} turns the fresh box evaluation (10{,}000 samples per model) into a certificate that needs no client data. The worst-run 95\% upper bound on the violated volume is at most 0.19\% for CEGIS-FL on every task, against up to 1.3\% for Grad-Box and 13\% for Grad-Data, at their larger selected $\lambda$ (\cref{tab:ub}). On COMPAS and ACS Income, Grad-Box's bound (0.12\% and 0.06\%) matches CEGIS-FL's. These bounds are per model; simultaneously over all 40 CEGIS-FL models (Bonferroni, overall level 95\%) the bound is at most 0.27\%.
A small violated \emph{volume} is not the same as no violation. We ran the exact verifier of \cref{sec:verification} (sampling falsifier, then branch-and-bound with 30\,s per rule) on the selected models of seeds 0--2 (51 rule instances per method).
MonoNet is certified on every rule, by construction; no unconstrained model is certified on \emph{any} rule, CEGIS-FL included. The falsifier finds autograd-verified counterexamples for 63\% of CEGIS-FL's rules, 71\% of Grad-Box's and 78\% of FedAvg's; branch-and-bound is inconclusive for the rest. On average it proves monotonicity on 25\% of the domain volume for CEGIS-FL and Grad-Box and 15\% for FedAvg.
The counterexamples found for CEGIS-FL have signed slopes of $-0.10$ to $-0.19$ (logit per standardised unit), so the remaining violations are small in volume but not negligible in magnitude where they occur.
Penalties therefore made violations \emph{rare}, not absent; only the architecture yielded a certificate. Larger budgets, tighter bounds or certification-oriented training \citep{liu2020certified} may change this.
The penalties cost one extra backward pass. Wall-clock time per run is $1.3$--$3.6\times$ that of FedAvg for the penalty methods and $1.4$--$1.6\times$ for MonoNet (\cref{tab:cost}).

\section{Discussion and limitations}
\label{sec:discussion}

\paragraph{What to use.}
If the architecture can be chosen, a monotone-by-construction network is the simplest option: the only one we could certify, preserved exactly by FedAvg (\cref{prop:convex}), tuning-free, and as accurate as unconstrained models on all six tasks.
If the architecture is fixed (an existing model, a required feature extractor, or rules that change after deployment), a gradient penalty on the clients' own data is enough to control violations \emph{where data lives}. If the model will also be queried on rare or out-of-range inputs, the penalty should also be evaluated on samples from the input domain, keeping the client term (box samples alone leave violations near the data). This is more $\lambda$-efficient over the domain, and since AUC stayed flat up to $\lambda=1000$, a large $\lambda$ is a reasonable default. Verified counterexamples (CEGIS-FL) reach volumes that \cref{prop:cp} bounds below 0.2\% at a much smaller $\lambda$; on German and Heart only CEGIS-FL reached them within our grid.
Random-pair losses and metrics should not be used to enforce or assess monotonicity.

\paragraph{Why not always use the architecture?}
Penalties apply to any differentiable model, rules can change without changing the architecture or be enforced by fine-tuning, and the same machinery audits models trained without the rule. We did not test larger or non-tabular models, where architectural constraints may cost more, nor lattice models or monotone gradient-boosted trees in FL.
Our server-side correction step \emph{increased} violations; a diagnostic traces this to the server optimiser (\cref{app:server}), so we draw no conclusion about where constraints belong.

\paragraph{The public input box.}
Methods that sample $\U(\X)$ rely on agreed bounds and category frequencies, which have a privacy cost (\cref{sec:setup}). Exact extremes can identify individuals and secure aggregation does not compute them, so bounds should come from a public schema or a differentially private release. A DP release typically \emph{widens} the box, moving more of the sampling budget and of $\Vol_\U$ to inputs far from any data. A box narrower than the true range would instead leave real inputs unpenalised. On the data-supported box (\cref{tab:clip}) the advantage of box samples over client data shrinks. The trade-off between privacy budget, box tightness and enforcement efficiency is open.

\paragraph{Limitations.}
\emph{(i) Scope.} Six tabular tasks, 3--10 clients with full participation, two-layer MLPs and FedAvg only; cross-device FL with sparse participation, high-dimensional or multimodal inputs and other server optimisers (FedProx, FedAvgM, FedAdam, SCAFFOLD) are untested. Uniform sampling finds violations less reliably in high dimensions, so larger problems need gradient- or solver-based falsifiers.
\emph{(ii) Ablations not run.} DP-noised boxes, CEGIS-FL beyond $\lambda=100$, and baselines such as lattices, monotone GBDTs and COMET. Our ablations (\cref{sec:res-ablation}) use two values of $\lambda$ and the in-run evaluation; on pooled data the location effect is consistent with centralized findings \citep{liu2020certified,monteiro2022regularization}.
\emph{(iii) Certification.} Exact verification was conclusive only for constrained networks within 30\,s per rule; for penalised networks we rely on per-model statistical bounds on the violated volume.
\emph{(iv) Rules.} A misspecified rule costs accuracy and encodes the misspecification (hence no age rule on Adult); some rules (age, and the direction of maximum heart rate, on Heart) are judgments that may differ between sites (\cref{app:datasets}).
\emph{(v) Statistics.} With five seeds the smallest two-sided Wilcoxon $p$ is 0.0625, hence sign counts and ten seeds on the small tasks.
\emph{(vi) Privacy.} We do not combine the penalties with differentially private training.

\paragraph{Broader impact.}
Monotonicity rules are often motivated by fairness and accountability, but a monotone model is not a fair model. On COMPAS, for example, prior offences may themselves reflect biased policing, and a rule that risk increases with priors encodes that assumption. We excluded race but proxies remain. Our contribution is a way to \emph{check} and \emph{enforce} stated rules faithfully. Whether a rule is appropriate is a domain decision that our methods do not make.

\section{Conclusion}
Monotonicity rules can be enforced in federated learning without measurable loss in AUC, and a server can audit them without client data, because monotonicity is a property of the function. A gradient penalty on client data controls violations where data lives. Evaluating it also on samples from the shared input domain, and on counterexamples found by the server, controls them over the domain far more efficiently; ablations show that where these points lie matters more than how many there are. Hard guarantees still require monotone architectures, which federated averaging preserves exactly. The random-pair loss and metric are poor substitutes for ceteris-paribus methods. Our code and per-run logs, available at \url{https://github.com/AshrafulKabir7/fed-monotone}, regenerate every number in this paper.

\paragraph{Acknowledgements.} This work grew out of a graduate course project at BRAC University.

\paragraph{Use of AI assistance.} The authors used an AI assistant for code development, experiment orchestration, data analysis and drafting of the manuscript. All results were produced by the released code, and all claims, proofs, numbers and references were checked by the authors, who take full responsibility for the content.

\bibliography{references}
\bibliographystyle{tmlr}

\appendix
\section{Proofs}
\label{app:proofs}

\paragraph{\Cref{prop:mvr}.}
$\partial_m f=a>0$, so $f$ is strictly increasing in $x_m$. Let $\Delta_m=X_m-X'_m$ and $\Delta_j=X_j-X'_j$. They are independent, and $\Delta_j$ is symmetric about 0 with a continuous distribution function $F$, so $F(0)=1/2$. For $b>0$,
\[
\mathrm{MVR}=\Pr[a\Delta_m+b\Delta_j<0\mid\Delta_m>0]=\mathbb{E}\big[F(-\tfrac{a}{b}\Delta_m)\,\big|\,\Delta_m>0\big].
\]
As $b/a\to\infty$, $F(-\tfrac{a}{b}\Delta_m)\to F(0)=1/2$ pointwise by continuity, and $0\le F\le 1$, so dominated convergence gives $\mathrm{MVR}\to1/2$.
For positivity, let $b>0$ and $X_j$ be non-degenerate. Then $\Delta_j$ is symmetric and not a.s.\ zero, so there is $\delta>0$ with $\Pr[\Delta_j<-\delta]>0$. For any $\varepsilon>0$, $\Pr[0<\Delta_m<\varepsilon]>0$: since $X_m$ is continuous, $\Pr[|X_m-X'_m|<\varepsilon]>0$ and $\Pr[\Delta_m=0]=0$, and $\Delta_m$ is symmetric. Take $\varepsilon=b\delta/a$. By independence, $\Pr[a\Delta_m+b\Delta_j<0,\ \Delta_m>0]\ge\Pr[0<\Delta_m<\varepsilon]\,\Pr[\Delta_j<-\delta]>0$. \qed

\paragraph{\Cref{prop:convex}.}
Convexity of $\Theta_{\mathcal M}$ means $\sum_k p_k\theta_k\in\Theta_{\mathcal M}$ for $\theta_k\in\Theta_{\mathcal M}$, $p_k\ge0$ and $\sum_kp_k=1$, and every element of $\Theta_{\mathcal M}$ gives a monotone function. FedAvg and FedProx aggregate by exactly such a weighted average. For a reparameterisation, $\Theta_{\mathcal M}=\Theta$. \qed

\paragraph{\Cref{ex:average}.}
$f_A'(x)=2\cdot\mathbb 1[x>0]\ge0$. $f_B(x)=-3(2-x)$ for $x<2$ and $0$ otherwise, so $f_B'(x)=3\cdot\mathbb 1[x<2]\ge0$. The average has $(w,b,v)=(\tfrac12,1,-1)$, so $f(x)=-\relu(x/2+1)$ with $f'(x)=-\tfrac12<0$ for $x>-2$. \qed

\paragraph{\Cref{prop:support}.}
\textbf{(a)} By linearity of expectation, $\sum_kp_k\,\mathbb{E}_{P_k}\rho=\mathbb{E}_{\sum_kp_kP_k}\rho=R_P$.
\textbf{(b)} If $x$ violates some rule at margin $\tau$, i.e.\ $-s_m\partial_mf(x)>\tau$, then $\rho(f,x)>\tau^2$. Markov's inequality gives $\mu\{\rho\ge\tau^2\}\le R_\mu(f)/\tau^2$.
\textbf{(c)} With $V$ the violation set, $\Vol_\U(f)=\int\mathbb 1_V\,d\U=\int\mathbb 1_V\,\tfrac{d\U}{dP}\,dP\le r\,P(V)$. For the second statement, $R_P(f)=\int\rho(f,x)\,dP(x)$ depends only on $\nabla f$ $P$-almost everywhere. Let $N\supseteq\operatorname{supp}P$ be open, and let $\chi$ be a smooth function that equals 0 on a neighbourhood of $\operatorname{supp}P$ inside $N$ and 1 outside $N$. Put $g=f-c\,s_m x_m\chi$. Then $g=f$ near $\operatorname{supp}P$, so $R_P(g)=R_P(f)$. Outside $N$, $s_m\partial_mg=s_m\partial_mf-c$, which is below $-\tau$ for $c$ large enough wherever $\partial_m f$ is bounded. Hence $\Vol_\U(g)\ge\U(\X\setminus N)$, which approaches $\U(\X\setminus\operatorname{supp}P)$ as $N$ shrinks to $\operatorname{supp}P$. \qed

\paragraph{\Cref{prop:cp}.}
Given $f$, the indicators $\mathbb 1[x_i\in V]$ are i.i.d.\ Bernoulli with mean $\Vol_\U(f)$, so $k\sim\mathrm{Bin}(N,\Vol_\U(f))$. The one-sided Clopper--Pearson bound $B^{-1}(1-\delta;k+1,N-k)$, the $(1-\delta)$-quantile of a $\mathrm{Beta}(k+1,N-k)$ distribution, has coverage at least $1-\delta$ for every value of the parameter \citep{clopper1934}. For $k=0$ it solves $(1-p)^N=\delta$. Independence of the samples from $f$ holds because the fresh evaluation samples are drawn with a seed that is used neither in training nor in selection. \qed

\section{Experimental details}
\label{app:details}

\subsection{Tasks and monotonicity rules}
\label{app:datasets}

\paragraph{Synthetic.}
$x_1,x_2\sim\mathcal U(-2,2)$, $x_3,x_4,x_5\sim\mathcal N(0,1)$, and $c\in\{a,b,c\}$ uniform with effects $(0,0.7,-0.7)$. The logit is $f^*(x)=0.25\,(1.2x_1+0.8\tanh 2x_2)+\sin 2x_3+0.6\,x_4x_5+\mathrm{eff}(c)$ and $y\sim\mathrm{Bernoulli}(\sigma(2f^*))$, with $n=20{,}000$.
$f^*$ is increasing in $x_1$ and $x_2$. The factor $0.25$ makes these effects weak relative to the non-monotone terms; with a factor of 1, unconstrained FedAvg already learns a monotone function (calibration, \cref{app:protocol}). The ten clients own consecutive quantile bands of $x_1$, so each client sees only a narrow range of a constrained feature.

\paragraph{Adult} (UCI census income, $n=48{,}842$; label: income $>\$50$K).
Numeric features: age, education-num, capital gain, capital loss, hours per week. Categorical features: workclass, marital status, occupation, relationship, race, sex, native country.
Rules ($+$): education-num, hours per week and capital gain, each of which, all else equal, should not lower the probability of high income. Age is not constrained (income is hump-shaped in age).

\paragraph{COMPAS} ($n=6{,}172$ after the standard ProPublica filtering: screening within $\pm30$ days of arrest, known recidivism status, no ordinary traffic offences; label: two-year recidivism).
Features: age, prior count, three juvenile counts, charge degree, sex. Race is excluded.
Rules ($+$): prior count and the juvenile felony, misdemeanour and other counts, as in \citet{liu2020certified}.

\paragraph{German credit} ($n=1{,}000$; label: good credit).
Seven numeric and thirteen categorical features.
Rules ($-$): loan duration and credit amount. All else equal, a longer or larger loan should not make a good rating more likely.

\paragraph{Heart disease} (four hospitals: Cleveland 303, Hungary 294, Switzerland 123, VA Long Beach 200; label: any disease).
Slope, number of major vessels and thal are dropped because they are mostly missing outside Cleveland. Cholesterol values of 0 (Switzerland) are treated as missing and imputed with the federated mean.
Rules: age, resting blood pressure and exercise-induced ST depression ($+$); maximum heart rate achieved ($-$, since lower exercise capacity indicates higher risk). Cholesterol is not constrained because its association is inconsistent across sites.

\paragraph{ACS Income} (folktables, 2018 1-year survey; label: income $>\$50$K).
States: CA, TX, NY, FL, IL, NM and MS (each subsampled to 8{,}000 people with a fixed seed), and WY, VT and ME (kept whole), $n=69{,}833$.
Features: age, educational attainment (ordinal), usual hours worked per week, class of worker, marital status, relationship, sex, race. Occupation and place of birth are dropped (hundreds of codes).
Rules ($+$): educational attainment and hours worked.

\subsection{Protocol and calibration}
\label{app:protocol}
\paragraph{Federated-safe preprocessing.} Each client sends counts, sums, sums of squares, minima, maxima and category counts of its \emph{training} split. The server forms global means and standard deviations, the box bounds and the category vocabularies and frequencies. The numeric transform equals a pooled standard scaler to $10^{-4}$ (unit test).
\paragraph{Partitions.} Dirichlet label skew resamples until every client has at least 20 examples (100 for German). Each client's train/validation/test split is stratified when both classes have at least three examples.
\paragraph{Calibration.} With one seed and 100 rounds, validation AUC of FedAvg peaked at round 10 on Adult, 90 on COMPAS, 75 on ACS Income, 50--85 on German, 45--70 on Heart and 95 on the original Synthetic task (monotone factor 1; the weakened version used in the paper peaked at round 100), while FedAvg's violated fraction kept increasing (e.g.\ Adult: 18\% at round 10, 81\% at round 100). We therefore fixed the round budgets in \cref{tab:datasets} and select checkpoints by validation AUC.

\subsection{Hyperparameters}
Box points per step: 64. Training margin inside $\rho$: 0 (i.e.\ $\rho$ penalises any negative signed slope). Evaluation tolerance for counting a violation: $\tau=10^{-4}$, with $0.01$ and $0.1$ in \cref{tab:magnitude}. Counterexample points per step (CEGIS-FL): up to 64.
CEGIS-FL: search every $T=5$ rounds, $N=20{,}000$ candidates, $c=8$ counterexamples per rule, $J=8$ jittered copies (2\% of each numeric range), pool capacity 2{,}048.
Server step (Grad-Box + server): 5 Adam steps with learning rate $10^{-3}$ on 512 box points, $\lambda_s=1$, $\mu_s=10$.
Pairwise: 1{,}024 random pairs per minibatch.
MonoNet: the same widths as the MLP; activations split in thirds (ReLU, $-\relu(-z)$, hardtanh).

\subsection{Verification and tests}
The MILP uses HiGHS \citep{huangfu2018highs} through SciPy. Branch-and-bound is implemented in PyTorch \citep{paszke2019pytorch} with batched interval arithmetic. Our test suite (17 tests) checks, among other things, that:
\begin{itemize}
\item a monotone linear model has zero ceteris-paribus violations but a random-pair MVR above 0.2;
\item MonoNet is monotone for random parameters and after FedAvg;
\item every counterexample returned by the falsifier or the MILP has a negative autograd derivative, matching the MILP objective to $10^{-4}$;
\item the MILP never certifies a rule that dense sampling shows to be violated;
\item branch-and-bound certifies a network that is monotone by construction, and its proven certified and violated volumes are consistent with Monte Carlo estimates;
\item federated preprocessing equals pooled preprocessing.
\end{itemize}

\subsection{Reproducibility}
Every run writes its full configuration, per-round history and selected checkpoint. Tables and figures are generated from these files by \texttt{scripts/paper\_figures.py}. The main sweep is \texttt{sweeps/main.yaml} with the extensions \texttt{main\_ext.yaml}, \texttt{main\_seeds.yaml} and \texttt{ext2\_lambda*.yaml} ($\lambda\in\{300,1000\}$); the ablations are \texttt{sweeps/ablations*.yaml} (tables by \texttt{scripts/ablation\_analysis.py}); the heterogeneity sweeps are \texttt{sweeps/heterogeneity*.yaml}. All experiments ran on a CPU-only workstation (8\,GB RAM, 4--6 parallel workers).

\section{Additional results}
\label{app:additional}

\begin{table}[h]
\caption{Validation-selected $\lambda$. Grids: Grad-Data and Grad-Box $\{0.1,1,10,30,100,300,1000\}$; CEGIS-FL $\{0.1,1,10,30,100\}$; Pairwise $\{0.01,0.1,1,10\}$; Grad-Mixup and Grad-Box + server $\{0.1,1,10\}$.}
\label{tab:lambda}
\centering\small
\begin{tabular}{lcccccc}
\toprule
Method & Synthetic & Adult & COMPAS & German & Heart & ACS Income \\
\midrule
Pairwise & 0.01 & 0.1 & 0.1 & 0.01 & 0.1 & 0.1 \\
Grad-Data & 1000 & 300 & 300 & 1000 & 1000 & 1000 \\
Grad-Box & 300 & 100 & 1000 & 1000 & 1000 & 1000 \\
CEGIS-FL & 10 & 1 & 30 & 100 & 100 & 100 \\
\bottomrule
\end{tabular}

\end{table}

\paragraph{Pairwise loss across $\lambda$.}
\Cref{tab:pairwise} lists, for each $\lambda$, the change in validation AUC relative to FedAvg and the violated fraction. Beyond $\lambda=0.1$ the loss costs 0.008--0.26 AUC, and its effect on violations is not monotone in $\lambda$.

\begin{table}[h]
\caption{Pairwise loss: $\Delta$ validation AUC vs.\ FedAvg / violated fraction of the box (\%), mean over seeds.}
\label{tab:pairwise}
\centering\small
\begin{tabular}{lcccccc}
\toprule
 & Synthetic & Adult & COMPAS & German & Heart & ACS Income \\
\midrule
$\lambda=0.01$ & +0.000 / 10.3 & +0.001 / 16.7 & +0.000 / 80.6 & +0.000 / 7.8 & +0.000 / 47.3 & +0.000 / 28.3 \\
$\lambda=0.1$ & -0.008 / 2.9 & -0.002 / 4.4 & -0.001 / 57.6 & -0.007 / 7.3 & -0.001 / 32.8 & -0.000 / 25.7 \\
$\lambda=1$ & -0.241 / 2.9 & -0.031 / 4.1 & -0.008 / 86.8 & -0.102 / 6.0 & -0.029 / 1.8 & -0.019 / 30.0 \\
$\lambda=10$ & -0.263 / 5.3 & -0.070 / 25.5 & -0.024 / 97.1 & -0.158 / 18.5 & -0.145 / 87.1 & -0.040 / 2.3 \\
\bottomrule
\end{tabular}

\end{table}

\paragraph{Variants that did not help.}
At matched $\lambda\in\{0.1,1,10\}$, adding mixup points to Grad-Data reduces violations in only 8 of 18 task--$\lambda$ settings, with a clear benefit only on Adult. Adding the server-side correction step to Grad-Box \emph{increases} violations in 17 of 18 settings, without affecting AUC; the diagnostic in \cref{app:server} traces this to the server optimiser. \Cref{tab:variants} reports both variants at their selected $\lambda$.

\begin{table}[h]
\caption{AUC difference to FedAvg (paired over seeds): mean [95\% CI]. $^\dagger$: equivalence within $\pm0.01$ not established (TOST, $p\ge0.05$); all other cells are equivalent.}
\label{tab:auc-ci}
\centering\scriptsize
\resizebox{\linewidth}{!}{\begin{tabular}{lcccccc}
\toprule
 & Synthetic & Adult & COMPAS & German & Heart & ACS Income \\
\midrule
Pairwise & +0.000 [-0.000, +0.000] & -0.003 [-0.007, +0.002] & +0.001 [-0.002, +0.004] & +0.002 [+0.000, +0.003] & -0.002 [-0.006, +0.002] & +0.000 [-0.001, +0.001] \\
Grad-Data & +0.000 [-0.000, +0.000] & -0.000 [-0.001, +0.000] & +0.001 [-0.003, +0.004] & +0.002 [-0.002, +0.006] & -0.000 [-0.004, +0.003] & -0.000 [-0.001, +0.000] \\
Grad-Box & +0.000 [-0.000, +0.001] & -0.001 [-0.002, +0.001] & +0.001 [-0.003, +0.005] & +0.005 [+0.001, +0.009] & -0.001 [-0.005, +0.002] & -0.000 [-0.001, +0.000] \\
CEGIS-FL & +0.001 [+0.000, +0.001] & -0.000 [-0.002, +0.002] & -0.000 [-0.007, +0.007] & +0.003 [-0.002, +0.008] & -0.001 [-0.005, +0.003] & -0.000 [-0.001, +0.000] \\
MonoNet & +0.001 [-0.001, +0.003] & -0.000 [-0.001, +0.001] & -0.000 [-0.005, +0.005] & -0.001 [-0.014, +0.012]$^\dagger$ & -0.000 [-0.006, +0.005] & -0.001 [-0.001, -0.000] \\
\bottomrule
\end{tabular}
}
\end{table}

\begin{table}[h]
\caption{$\lambda$-efficiency: factor by which Grad-Data's $\lambda$ must exceed Grad-Box's ($\lambda=10$) to reach the same violated fraction of the box. Log--log interpolation over the $\lambda$ grid; [95\% seed-bootstrap CI]. ``$>100$'': Grad-Data does not reach the level within the grid ($\lambda\le1000$).}
\label{tab:shift}
\centering\scriptsize
\resizebox{\linewidth}{!}{\begin{tabular}{lcccccc}
\toprule
 & Synthetic & Adult & COMPAS & German & Heart & ACS Income \\
\midrule
$\lambda_{\mathrm{Data}}/\lambda_{\mathrm{Box}}$ & 2.5 [1.1, 5.1] & 14.9 [2.4, $>\!100$] & $>\!100$ [1.5, $\infty$) & 6.0 [0.6, $>\!100$] & 3.9 [0.9, 17.4] & 9.8 [5.6, 24.4] \\
\bottomrule
\end{tabular}
}
\end{table}

\begin{table}[h]
\caption{Ablations (\cref{sec:res-ablation}): violated fraction (\%) of the box ($\Vol_\U$) and of test points ($\Vol_P$) at matched $\lambda$, mean over seeds (5; 10 on German and Heart), in-run evaluation. $B$: the client's minibatch; ``64 box'' and ``64 data'': additional penalty points per step from the box or from the client's own data.}
\label{tab:ablation}
\centering\scriptsize
\resizebox{\linewidth}{!}{\begin{tabular}{lcccccc}
\toprule
 & Synthetic & Adult & COMPAS & German & Heart & ACS Income \\
\midrule
Grad-Data ($B$), $\Vol_\U$, $\lambda=10$ & 1.08 & 6.40 & 13.41 & 4.78 & 16.02 & 7.90 \\
Grad-Data$\times$2 ($B$ + 64 data), $\Vol_\U$, $\lambda=10$ & 1.15 & 2.03 & 19.25 & 4.50 & 16.45 & 7.78 \\
Grad-Box ($B$ + 64 box), $\Vol_\U$, $\lambda=10$ & 0.61 & 0.10 & 1.60 & 2.23 & 10.59 & 1.34 \\
Box-only (64 box), $\Vol_\U$, $\lambda=10$ & 0.54 & 0.07 & 1.24 & 2.17 & 10.38 & 0.87 \\
\midrule
Grad-Data ($B$), $\Vol_P$, $\lambda=10$ & 0.82 & 0.83 & 6.48 & 1.29 & 10.38 & 0.51 \\
Grad-Data$\times$2 ($B$ + 64 data), $\Vol_P$, $\lambda=10$ & 0.90 & 0.54 & 7.91 & 1.69 & 11.24 & 0.56 \\
Grad-Box ($B$ + 64 box), $\Vol_P$, $\lambda=10$ & 0.81 & 0.60 & 8.33 & 1.04 & 13.35 & 0.34 \\
Box-only (64 box), $\Vol_P$, $\lambda=10$ & 1.62 & 1.53 & 31.87 & 1.44 & 17.35 & 0.44 \\
\midrule
Grad-Data ($B$), $\Vol_\U$, $\lambda=100$ & 0.23 & 0.20 & 4.22 & 1.77 & 8.50 & 1.31 \\
Grad-Data$\times$2 ($B$ + 64 data), $\Vol_\U$, $\lambda=100$ & 0.35 & 0.05 & 2.41 & 1.80 & 8.23 & 1.35 \\
Grad-Box ($B$ + 64 box), $\Vol_\U$, $\lambda=100$ & 0.10 & 0.00 & 0.24 & 0.77 & 2.90 & 0.14 \\
Box-only (64 box), $\Vol_\U$, $\lambda=100$ & 0.12 & 0.03 & 0.13 & 0.55 & 3.02 & 0.13 \\
\midrule
Grad-Data ($B$), $\Vol_P$, $\lambda=100$ & 0.16 & 0.08 & 1.67 & 0.55 & 2.97 & 0.05 \\
Grad-Data$\times$2 ($B$ + 64 data), $\Vol_P$, $\lambda=100$ & 0.21 & 0.04 & 2.67 & 0.74 & 3.62 & 0.06 \\
Grad-Box ($B$ + 64 box), $\Vol_P$, $\lambda=100$ & 0.18 & 0.03 & 2.34 & 0.30 & 3.95 & 0.06 \\
Box-only (64 box), $\Vol_P$, $\lambda=100$ & 0.38 & 0.55 & 22.09 & 0.25 & 5.30 & 0.08 \\
\bottomrule
\end{tabular}
}
\end{table}

\begin{table}[h]
\caption{Penalties trained on pooled data (``central'', one client) vs.\ federated, $\lambda=100$: violated fraction of the box (\%), in-run evaluation, mean over seeds. Last row: centralized FedAvg.}
\label{tab:central}
\centering\scriptsize
\resizebox{\linewidth}{!}{\begin{tabular}{lcccccc}
\toprule
 & Synthetic & Adult & COMPAS & German & Heart & ACS Income \\
\midrule
Grad-Data ($B$) (fed.) & 0.23 & 0.20 & 4.22 & 1.77 & 8.50 & 1.31 \\
Grad-Box ($B$ + 64 box) (fed.) & 0.10 & 0.00 & 0.24 & 0.77 & 2.90 & 0.14 \\
CEGIS-FL (fed.) & 0.00 & 0.00 & 0.02 & 0.03 & 0.04 & 0.01 \\
\midrule
Grad-Data ($B$) (central) & 0.17 & 0.14 & 10.38 & 1.25 & 10.88 & 2.38 \\
Grad-Box ($B$ + 64 box) (central) & 0.25 & 0.13 & 0.75 & 0.83 & 2.54 & 0.15 \\
CEGIS-FL (central) & 0.08 & 0.05 & 0.02 & 0.02 & 0.53 & 0.56 \\
\midrule
FedAvg (central) & 21.74 & 47.90 & 82.86 & 7.48 & 64.36 & 25.50 \\
\bottomrule
\end{tabular}
}
\end{table}

\begin{table}[h]
\caption{Number of runs per task (main experiment). The heterogeneity study (120 runs) and the ablations (280 runs) are separate.}
\label{tab:runs}
\centering\small
\begin{tabular}{lrrrr}
\toprule
Task & runs & federated & centralized & seeds \\
\midrule
Synthetic & 165 & 155 & 10 & 5 \\
Adult & 165 & 155 & 10 & 5 \\
COMPAS & 165 & 155 & 10 & 5 \\
German & 330 & 310 & 20 & 10 \\
Heart & 330 & 310 & 20 & 10 \\
ACS Income & 165 & 155 & 10 & 5 \\
\midrule
Total & 1320 & 1240 & 80 &  \\
\bottomrule
\end{tabular}

\end{table}

\begin{figure}[h]
\centering
\includegraphics[width=\linewidth]{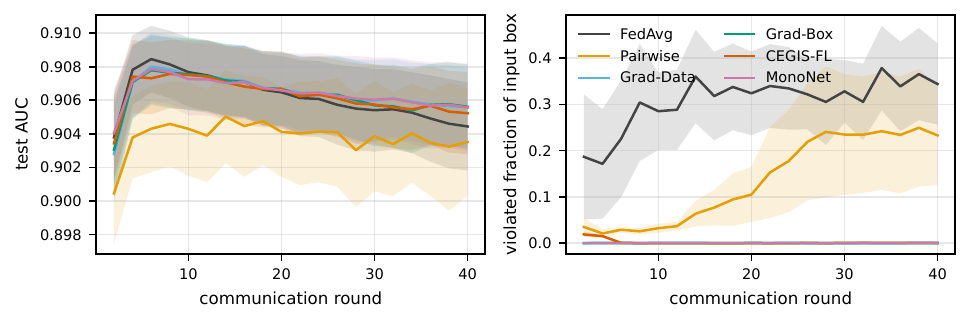}
\caption{Adult, Dirichlet $\alpha=0.3$, 10 clients: test AUC (left) and violated fraction of the input box (right) over communication rounds, mean $\pm$ s.e.\ over 5 seeds at the selected $\lambda$. FedAvg's violations grow while AUC declines. Pairwise slows the growth but does not stop it.}
\label{fig:convergence}
\end{figure}

\begin{figure}[h]
\centering
\includegraphics[width=\linewidth]{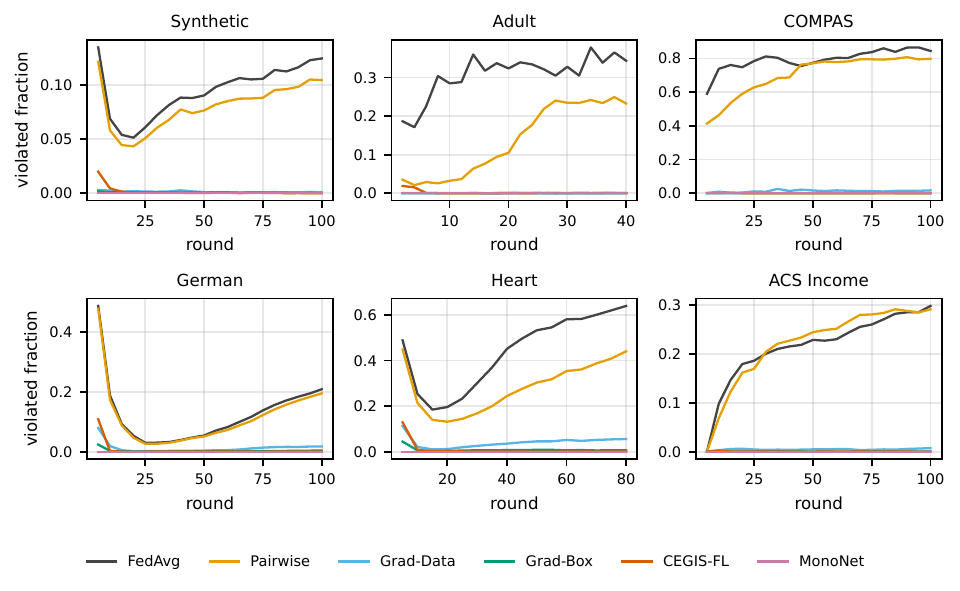}
\caption{Violated fraction of the input box over communication rounds on all tasks (mean over seeds, selected $\lambda$). On every task, after the first few rounds (dominated by the random initialisation), FedAvg's violations increase with further training. The pairwise loss follows the same trend at a lower level, while the gradient penalties and MonoNet stay near zero.}
\label{fig:convergence-all}
\end{figure}

\begin{table}[h]
\caption{Violated fraction (\%) of the variants at their selected $\lambda$, next to the methods they extend. The grids differ ($\le10$ for the variants, $\le1000$ for Grad-Data and Grad-Box), so this table is not a like-for-like comparison; the matched-$\lambda$ comparison is in the text.}
\label{tab:variants}
\centering\small
\begin{tabular}{lcccccc}
\toprule
 & Synthetic & Adult & COMPAS & German & Heart & ACS Income \\
\midrule
Grad-Data & 0.1{\scriptsize$\pm$0.0} & 0.0{\scriptsize$\pm$0.1} & 0.5{\scriptsize$\pm$0.8} & 1.3{\scriptsize$\pm$2.0} & 4.8{\scriptsize$\pm$4.0} & 0.5{\scriptsize$\pm$0.3} \\
Grad-Mixup & 1.5{\scriptsize$\pm$0.5} & 3.6{\scriptsize$\pm$3.9} & 13.8{\scriptsize$\pm$19.0} & 5.8{\scriptsize$\pm$6.1} & 16.2{\scriptsize$\pm$9.5} & 7.8{\scriptsize$\pm$3.0} \\
Grad-Box & 0.0{\scriptsize$\pm$0.0} & 0.0{\scriptsize$\pm$0.0} & 0.0{\scriptsize$\pm$0.0} & 0.3{\scriptsize$\pm$0.3} & 0.6{\scriptsize$\pm$0.2} & 0.0{\scriptsize$\pm$0.0} \\
Grad-Box + server & 1.5{\scriptsize$\pm$0.2} & 0.1{\scriptsize$\pm$0.1} & 2.5{\scriptsize$\pm$1.7} & 8.4{\scriptsize$\pm$3.6} & 19.2{\scriptsize$\pm$5.9} & 2.5{\scriptsize$\pm$1.1} \\
\bottomrule
\end{tabular}

\end{table}

\begin{table}[h]
\caption{Test AUC (mean $\pm$ s.d.) for the runs in \cref{tab:violation}. All enforcement methods stay within $\pm0.005$ of FedAvg.}
\label{tab:auc}
\centering
\small
\begin{tabular}{lcccccc}
\toprule
Method & Synthetic & Adult & COMPAS & German & Heart & ACS Income \\
\midrule
FedAvg & 0.879{\scriptsize$\pm$0.002} & 0.908{\scriptsize$\pm$0.004} & 0.727{\scriptsize$\pm$0.011} & 0.772{\scriptsize$\pm$0.046} & 0.873{\scriptsize$\pm$0.019} & 0.854{\scriptsize$\pm$0.004} \\
Pairwise & 0.879{\scriptsize$\pm$0.002} & 0.906{\scriptsize$\pm$0.006} & 0.728{\scriptsize$\pm$0.011} & 0.774{\scriptsize$\pm$0.046} & 0.871{\scriptsize$\pm$0.020} & 0.855{\scriptsize$\pm$0.005} \\
Grad-Data & 0.879{\scriptsize$\pm$0.002} & 0.908{\scriptsize$\pm$0.005} & 0.728{\scriptsize$\pm$0.010} & 0.774{\scriptsize$\pm$0.045} & 0.873{\scriptsize$\pm$0.018} & 0.854{\scriptsize$\pm$0.005} \\
Grad-Box & 0.880{\scriptsize$\pm$0.002} & 0.907{\scriptsize$\pm$0.005} & 0.728{\scriptsize$\pm$0.010} & 0.777{\scriptsize$\pm$0.046} & 0.872{\scriptsize$\pm$0.019} & 0.854{\scriptsize$\pm$0.005} \\
CEGIS-FL & 0.880{\scriptsize$\pm$0.002} & 0.908{\scriptsize$\pm$0.005} & 0.727{\scriptsize$\pm$0.014} & 0.775{\scriptsize$\pm$0.046} & 0.872{\scriptsize$\pm$0.019} & 0.854{\scriptsize$\pm$0.005} \\
\midrule
MonoNet & 0.880{\scriptsize$\pm$0.003} & 0.908{\scriptsize$\pm$0.004} & 0.727{\scriptsize$\pm$0.009} & 0.771{\scriptsize$\pm$0.053} & 0.873{\scriptsize$\pm$0.021} & 0.854{\scriptsize$\pm$0.005} \\
\midrule
FedAvg (centralized) & 0.876{\scriptsize$\pm$0.006} & 0.914{\scriptsize$\pm$0.003} & 0.742{\scriptsize$\pm$0.006} & 0.778{\scriptsize$\pm$0.040} & 0.875{\scriptsize$\pm$0.027} & 0.855{\scriptsize$\pm$0.004} \\
MonoNet (centralized) & 0.879{\scriptsize$\pm$0.004} & 0.914{\scriptsize$\pm$0.003} & 0.739{\scriptsize$\pm$0.006} & 0.779{\scriptsize$\pm$0.049} & 0.878{\scriptsize$\pm$0.021} & 0.855{\scriptsize$\pm$0.004} \\
\bottomrule
\end{tabular}

\end{table}

\begin{table}[h]
\caption{Violated fraction (\%) of the \emph{data-supported box} $\Vol_{\U_q}$ (numeric ranges clipped to the 1st--99th test percentiles; 10{,}000 fresh samples), selected $\lambda$, mean $\pm$ s.d.\ over seeds.}
\label{tab:clip}
\centering\scriptsize
\begin{tabular}{lcccccc}
\toprule
 & Synthetic & Adult & COMPAS & German & Heart & ACS Income \\
\midrule
FedAvg & 11.18{\scriptsize$\pm$2.22} & 15.14{\scriptsize$\pm$6.11} & 85.33{\scriptsize$\pm$15.59} & 4.00{\scriptsize$\pm$3.86} & 44.24{\scriptsize$\pm$27.36} & 15.86{\scriptsize$\pm$3.13} \\
Pairwise & 9.56{\scriptsize$\pm$2.52} & 13.34{\scriptsize$\pm$6.95} & 73.57{\scriptsize$\pm$26.58} & 4.75{\scriptsize$\pm$5.47} & 23.47{\scriptsize$\pm$16.60} & 14.46{\scriptsize$\pm$3.53} \\
Grad-Data & 0.07{\scriptsize$\pm$0.03} & 0.03{\scriptsize$\pm$0.04} & 5.94{\scriptsize$\pm$6.40} & 0.81{\scriptsize$\pm$1.47} & 1.40{\scriptsize$\pm$0.65} & 0.15{\scriptsize$\pm$0.11} \\
Grad-Box & 0.08{\scriptsize$\pm$0.03} & 0.04{\scriptsize$\pm$0.08} & 2.06{\scriptsize$\pm$1.95} & 0.23{\scriptsize$\pm$0.32} & 0.61{\scriptsize$\pm$0.23} & 0.01{\scriptsize$\pm$0.01} \\
CEGIS-FL & 0.03{\scriptsize$\pm$0.01} & 0.04{\scriptsize$\pm$0.08} & 1.81{\scriptsize$\pm$2.70} & 0.01{\scriptsize$\pm$0.01} & 0.12{\scriptsize$\pm$0.07} & 0.01{\scriptsize$\pm$0.01} \\
MonoNet & 0.00{\scriptsize$\pm$0.00} & 0.00{\scriptsize$\pm$0.00} & 0.00{\scriptsize$\pm$0.00} & 0.00{\scriptsize$\pm$0.00} & 0.00{\scriptsize$\pm$0.00} & 0.00{\scriptsize$\pm$0.00} \\
\bottomrule
\end{tabular}

\end{table}

\paragraph{Paired tests at the selected $\lambda$.}
On fresh box samples, Grad-Box beats Grad-Data on Heart (9/10 seeds, Wilcoxon $p=0.006$), German (8/10, $p=0.035$), ACS Income and COMPAS (5/5 each, $p=0.062$, the minimum with five seeds), Synthetic (4/5) and Adult (3/5, two ties; both below 0.05\%). CEGIS-FL beats Grad-Box on Heart (10/10, $p=0.002$) and German (8/10, one tie, $p=0.008$) and on 4/5 seeds on Synthetic; on Adult, COMPAS and ACS Income most seeds are tied and both are at or below 0.02\%.

\begin{figure}[h]
\centering
\includegraphics[width=\linewidth]{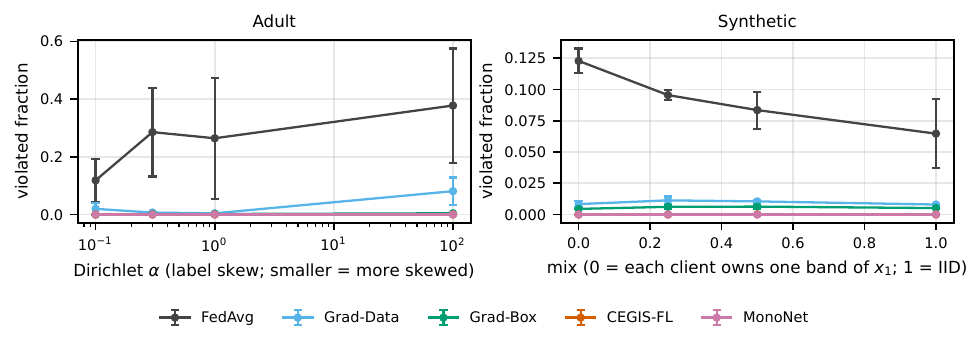}
\caption{Violated fraction against heterogeneity at matched $\lambda=10$ (mean $\pm$ s.e., 3 seeds). Left: Adult with Dirichlet label skew ($\alpha=100$ is close to IID). Right: Synthetic with covariate skew on the constrained feature $x_1$, interpolated towards IID by reassigning a fraction ``mix'' of examples at random.}
\label{fig:het}
\end{figure}

\paragraph{Heterogeneity (details).}
We vary heterogeneity at a fixed $\lambda=10$ for all penalty methods, with 3 seeds (\cref{fig:het}).
\emph{Covariate skew on a constrained feature} increases FedAvg's violations: on Synthetic, from 6.5\% when clients are IID to 12.3\% when each client owns one band of $x_1$.
\emph{Label skew} has no consistent effect on Adult. FedAvg's violated fraction varies between 12\% and 38\% across $\alpha$ with large seed-to-seed variation, and is highest in the near-IID setting. This is consistent with \cref{sec:res-baseline}: more effective training, not federation as such, is what drives violations.
Across all eight settings, the box-sample methods are unaffected by heterogeneity: Grad-Box stays at or below 0.6\%, and CEGIS-FL and MonoNet are at 0, with AUC equal to FedAvg's. Grad-Data stays at about 1\% on Synthetic at every level of skew, and on Adult ranges from 0.4\% to 8.1\%.
We therefore do \emph{not} find that client-data penalties degrade with skew, which \cref{prop:support} would allow. The domain-level advantage of box points is present at every level of heterogeneity we tested, IID included. It is a property of where the penalty is evaluated, not a symptom of heterogeneity.

\paragraph{Violation magnitude.}
$\Vol_\U$ uses a small tolerance ($\tau=10^{-4}$). \Cref{tab:magnitude} repeats the fresh evaluation with $\tau=0.01$ and $\tau=0.1$ (logit per standard deviation of the feature). FedAvg's violations persist at the stricter tolerances on every task except German, while all gradient penalties are at or near zero.

\begin{table}[h]
\caption{Violated fraction of the box (\%) at stricter tolerances $\tau$ (10{,}000 fresh samples, selected $\lambda$). MonoNet is 0 everywhere.}
\label{tab:magnitude}
\centering
\resizebox{\linewidth}{!}{\begin{tabular}{lcccccc}
\toprule
 & Synthetic & Adult & COMPAS & German & Heart & ACS Income \\
\midrule
FedAvg ($\tau=0.01$) & 10.99{\scriptsize$\pm$2.47} & 23.18{\scriptsize$\pm$21.64} & 83.82{\scriptsize$\pm$18.65} & 5.57{\scriptsize$\pm$6.60} & 47.12{\scriptsize$\pm$27.73} & 24.24{\scriptsize$\pm$5.36} \\
Pairwise ($\tau=0.01$) & 9.37{\scriptsize$\pm$2.42} & 4.09{\scriptsize$\pm$2.31} & 51.09{\scriptsize$\pm$37.54} & 6.34{\scriptsize$\pm$8.93} & 29.50{\scriptsize$\pm$20.05} & 24.02{\scriptsize$\pm$4.95} \\
Grad-Data ($\tau=0.01$) & 0.07{\scriptsize$\pm$0.02} & 0.01{\scriptsize$\pm$0.02} & 0.25{\scriptsize$\pm$0.39} & 0.87{\scriptsize$\pm$1.69} & 3.68{\scriptsize$\pm$3.34} & 0.34{\scriptsize$\pm$0.16} \\
Grad-Box ($\tau=0.01$) & 0.03{\scriptsize$\pm$0.02} & 0.00{\scriptsize$\pm$0.00} & 0.00{\scriptsize$\pm$0.00} & 0.14{\scriptsize$\pm$0.26} & 0.36{\scriptsize$\pm$0.19} & 0.01{\scriptsize$\pm$0.01} \\
CEGIS-FL ($\tau=0.01$) & 0.01{\scriptsize$\pm$0.01} & 0.00{\scriptsize$\pm$0.00} & 0.01{\scriptsize$\pm$0.01} & 0.00{\scriptsize$\pm$0.01} & 0.03{\scriptsize$\pm$0.01} & 0.00{\scriptsize$\pm$0.00} \\
\midrule
FedAvg ($\tau=0.1$) & 5.30{\scriptsize$\pm$1.86} & 3.92{\scriptsize$\pm$2.24} & 46.19{\scriptsize$\pm$36.36} & 0.35{\scriptsize$\pm$0.97} & 26.52{\scriptsize$\pm$23.45} & 10.27{\scriptsize$\pm$5.87} \\
Pairwise ($\tau=0.1$) & 4.33{\scriptsize$\pm$1.95} & 1.17{\scriptsize$\pm$0.77} & 30.15{\scriptsize$\pm$40.24} & 0.34{\scriptsize$\pm$0.57} & 10.45{\scriptsize$\pm$14.72} & 9.11{\scriptsize$\pm$6.05} \\
Grad-Data ($\tau=0.1$) & 0.00{\scriptsize$\pm$0.01} & 0.00{\scriptsize$\pm$0.00} & 0.00{\scriptsize$\pm$0.00} & 0.08{\scriptsize$\pm$0.25} & 0.27{\scriptsize$\pm$0.55} & 0.00{\scriptsize$\pm$0.00} \\
Grad-Box ($\tau=0.1$) & 0.00{\scriptsize$\pm$0.00} & 0.00{\scriptsize$\pm$0.00} & 0.00{\scriptsize$\pm$0.00} & 0.01{\scriptsize$\pm$0.04} & 0.00{\scriptsize$\pm$0.00} & 0.00{\scriptsize$\pm$0.00} \\
CEGIS-FL ($\tau=0.1$) & 0.00{\scriptsize$\pm$0.00} & 0.00{\scriptsize$\pm$0.00} & 0.00{\scriptsize$\pm$0.00} & 0.00{\scriptsize$\pm$0.00} & 0.00{\scriptsize$\pm$0.00} & 0.00{\scriptsize$\pm$0.00} \\
\bottomrule
\end{tabular}
}
\end{table}

\begin{figure}[h]
\centering
\includegraphics[width=\linewidth]{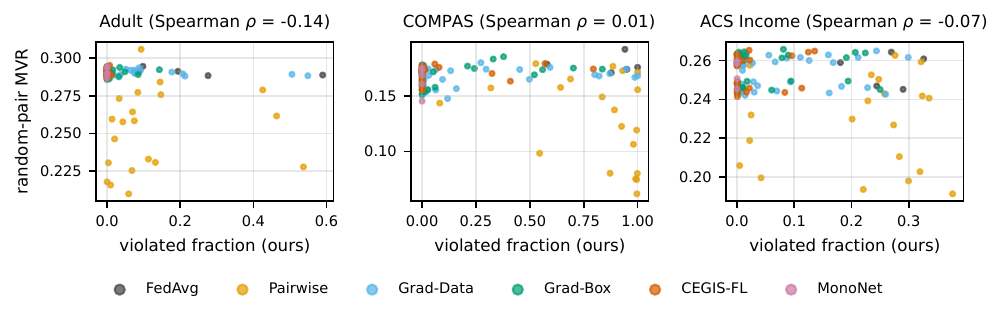}
\caption{Random-pair MVR against the ceteris-paribus violated fraction, one point per run. The random-pair metric does not track violations. MonoNet (0 violations by construction) scores like FedAvg, and the pairwise runs with the \emph{most} violations score the \emph{best}.}
\label{fig:mvr}
\end{figure}

\paragraph{Threshold metrics.}
At threshold 0.5, MonoNet has a lower F1 than FedAvg on COMPAS (0.41 vs.\ 0.51), with a lower predicted-positive rate; CEGIS-FL is at 0.45. A validation-tuned threshold would be the appropriate fix, which we did not evaluate; we flag that a monotone model's operating point can shift. \Cref{tab:threshold} gives balanced accuracy and F1 at threshold 0.5. Under label skew they vary strongly between seeds for every method, including FedAvg, which is why AUC is our primary accuracy metric.

\begin{table}[h]
\caption{Balanced accuracy and F1 at threshold 0.5 (mean $\pm$ s.d.).}
\label{tab:threshold}
\centering\scriptsize
\resizebox{\linewidth}{!}{\begin{tabular}{lcccccc}
\toprule
 & Synthetic & Adult & COMPAS & German & Heart & ACS Income \\
\midrule
FedAvg (bal.\ acc.) & 0.795{\scriptsize$\pm$0.004} & 0.684{\scriptsize$\pm$0.072} & 0.637{\scriptsize$\pm$0.053} & 0.603{\scriptsize$\pm$0.064} & 0.797{\scriptsize$\pm$0.033} & 0.752{\scriptsize$\pm$0.004} \\
Pairwise (bal.\ acc.) & 0.794{\scriptsize$\pm$0.005} & 0.670{\scriptsize$\pm$0.083} & 0.633{\scriptsize$\pm$0.046} & 0.619{\scriptsize$\pm$0.064} & 0.791{\scriptsize$\pm$0.033} & 0.746{\scriptsize$\pm$0.006} \\
Grad-Data (bal.\ acc.) & 0.794{\scriptsize$\pm$0.001} & 0.687{\scriptsize$\pm$0.067} & 0.634{\scriptsize$\pm$0.049} & 0.622{\scriptsize$\pm$0.067} & 0.797{\scriptsize$\pm$0.032} & 0.753{\scriptsize$\pm$0.004} \\
Grad-Box (bal.\ acc.) & 0.793{\scriptsize$\pm$0.003} & 0.688{\scriptsize$\pm$0.070} & 0.636{\scriptsize$\pm$0.042} & 0.619{\scriptsize$\pm$0.064} & 0.796{\scriptsize$\pm$0.030} & 0.752{\scriptsize$\pm$0.006} \\
CEGIS-FL (bal.\ acc.) & 0.794{\scriptsize$\pm$0.003} & 0.688{\scriptsize$\pm$0.071} & 0.616{\scriptsize$\pm$0.062} & 0.622{\scriptsize$\pm$0.065} & 0.792{\scriptsize$\pm$0.033} & 0.752{\scriptsize$\pm$0.006} \\
MonoNet (bal.\ acc.) & 0.794{\scriptsize$\pm$0.004} & 0.690{\scriptsize$\pm$0.057} & 0.605{\scriptsize$\pm$0.065} & 0.633{\scriptsize$\pm$0.063} & 0.801{\scriptsize$\pm$0.034} & 0.751{\scriptsize$\pm$0.005} \\
\midrule
FedAvg (F1) & 0.795{\scriptsize$\pm$0.005} & 0.515{\scriptsize$\pm$0.157} & 0.510{\scriptsize$\pm$0.190} & 0.834{\scriptsize$\pm$0.013} & 0.818{\scriptsize$\pm$0.029} & 0.677{\scriptsize$\pm$0.005} \\
Pairwise (F1) & 0.794{\scriptsize$\pm$0.005} & 0.481{\scriptsize$\pm$0.187} & 0.506{\scriptsize$\pm$0.150} & 0.838{\scriptsize$\pm$0.013} & 0.811{\scriptsize$\pm$0.029} & 0.667{\scriptsize$\pm$0.009} \\
Grad-Data (F1) & 0.793{\scriptsize$\pm$0.002} & 0.524{\scriptsize$\pm$0.141} & 0.498{\scriptsize$\pm$0.159} & 0.835{\scriptsize$\pm$0.013} & 0.816{\scriptsize$\pm$0.030} & 0.678{\scriptsize$\pm$0.006} \\
Grad-Box (F1) & 0.792{\scriptsize$\pm$0.004} & 0.524{\scriptsize$\pm$0.146} & 0.510{\scriptsize$\pm$0.135} & 0.836{\scriptsize$\pm$0.014} & 0.815{\scriptsize$\pm$0.026} & 0.676{\scriptsize$\pm$0.008} \\
CEGIS-FL (F1) & 0.793{\scriptsize$\pm$0.005} & 0.524{\scriptsize$\pm$0.149} & 0.447{\scriptsize$\pm$0.203} & 0.837{\scriptsize$\pm$0.016} & 0.814{\scriptsize$\pm$0.026} & 0.676{\scriptsize$\pm$0.008} \\
MonoNet (F1) & 0.793{\scriptsize$\pm$0.003} & 0.535{\scriptsize$\pm$0.124} & 0.408{\scriptsize$\pm$0.237} & 0.837{\scriptsize$\pm$0.017} & 0.821{\scriptsize$\pm$0.031} & 0.675{\scriptsize$\pm$0.007} \\
\bottomrule
\end{tabular}
}
\end{table}

\paragraph{Last-round checkpoints.}
Selecting the checkpoint by validation AUC favours FedAvg, whose violations grow with training. \Cref{tab:lastround} reports the last-round model instead. FedAvg's violations are larger (e.g.\ 34\% vs.\ 26\% on Adult, 21\% vs.\ 7\% on German, 64\% vs.\ 50\% on Heart). The ordering of the other methods is unchanged (on Adult, where Grad-Data, Grad-Box and CEGIS-FL are all at or below 0.1\%, the differences are within rounding).

\begin{table}[h]
\caption{Violated fraction (\%) of the last-round model.}
\label{tab:lastround}
\centering\small
\begin{tabular}{lcccccc}
\toprule
 & Synthetic & Adult & COMPAS & German & Heart & ACS Income \\
\midrule
FedAvg & 12.5{\scriptsize$\pm$2.3} & 34.4{\scriptsize$\pm$19.7} & 84.4{\scriptsize$\pm$19.2} & 20.9{\scriptsize$\pm$19.9} & 64.1{\scriptsize$\pm$14.6} & 29.8{\scriptsize$\pm$7.8} \\
Pairwise & 10.5{\scriptsize$\pm$2.1} & 23.3{\scriptsize$\pm$23.9} & 79.8{\scriptsize$\pm$27.2} & 19.5{\scriptsize$\pm$17.2} & 44.1{\scriptsize$\pm$16.9} & 29.1{\scriptsize$\pm$6.5} \\
Grad-Data & 0.1{\scriptsize$\pm$0.1} & 0.0{\scriptsize$\pm$0.0} & 1.7{\scriptsize$\pm$2.1} & 1.8{\scriptsize$\pm$2.1} & 5.6{\scriptsize$\pm$3.9} & 0.7{\scriptsize$\pm$0.4} \\
Grad-Box & 0.0{\scriptsize$\pm$0.0} & 0.0{\scriptsize$\pm$0.0} & 0.0{\scriptsize$\pm$0.0} & 0.4{\scriptsize$\pm$0.2} & 0.7{\scriptsize$\pm$0.3} & 0.0{\scriptsize$\pm$0.0} \\
CEGIS-FL & 0.0{\scriptsize$\pm$0.0} & 0.1{\scriptsize$\pm$0.1} & 0.0{\scriptsize$\pm$0.0} & 0.0{\scriptsize$\pm$0.0} & 0.1{\scriptsize$\pm$0.1} & 0.0{\scriptsize$\pm$0.0} \\
MonoNet & 0.0{\scriptsize$\pm$0.0} & 0.0{\scriptsize$\pm$0.0} & 0.0{\scriptsize$\pm$0.0} & 0.0{\scriptsize$\pm$0.0} & 0.0{\scriptsize$\pm$0.0} & 0.0{\scriptsize$\pm$0.0} \\
\bottomrule
\end{tabular}

\end{table}

\paragraph{Certificates and cost.}
\Cref{tab:ub} gives the worst-run 95\% Clopper--Pearson upper bound on the violated volume, and \cref{tab:cost} the wall-clock time relative to FedAvg. Wall-clock times come from runs that shared the machine with other runs, so differences below about 20\% are not meaningful.
The fresh evaluation agrees with the in-run metric used for $\lambda$ selection (Pearson $r=0.9995$, mean absolute difference 0.33 percentage points), so selection on the in-run metric did not bias the reported violations.
The finite-difference violation rate (step of 5\% of the range) agrees with the gradient-based rate across all selected federated runs (Spearman 0.998 on the fresh evaluation).

\begin{table}[h]
\caption{Worst-run one-sided 95\% upper bound (\%) on the violated volume (\cref{prop:cp}).}
\label{tab:ub}
\centering\small
\begin{tabular}{lcccccc}
\toprule
Method & Synthetic & Adult & COMPAS & German & Heart & ACS Income \\
\midrule
Grad-Data & 0.21 & 0.18 & 2.24 & 7.09 & 13.04 & 1.00 \\
Grad-Box & 0.16 & 0.05 & 0.12 & 1.32 & 1.21 & 0.06 \\
CEGIS-FL & 0.08 & 0.09 & 0.14 & 0.11 & 0.19 & 0.06 \\
\bottomrule
\end{tabular}

\end{table}

\begin{table}[h]
\caption{Wall-clock time per run relative to FedAvg; last row: FedAvg seconds per run.}
\label{tab:cost}
\centering\small
\begin{tabular}{lcccccc}
\toprule
Method & Synthetic & Adult & COMPAS & German & Heart & ACS Income \\
\midrule
FedAvg & 1.0$\times$ & 1.0$\times$ & 1.0$\times$ & 1.0$\times$ & 1.0$\times$ & 1.0$\times$ \\
Pairwise & 2.4$\times$ & 1.9$\times$ & 2.1$\times$ & 1.3$\times$ & 2.1$\times$ & 2.7$\times$ \\
Grad-Data & 2.3$\times$ & 2.6$\times$ & 2.3$\times$ & 1.4$\times$ & 1.5$\times$ & 2.1$\times$ \\
Grad-Box & 2.2$\times$ & 3.2$\times$ & 3.5$\times$ & 1.5$\times$ & 1.5$\times$ & 2.8$\times$ \\
CEGIS-FL & 3.1$\times$ & 3.0$\times$ & 3.6$\times$ & 2.2$\times$ & 1.6$\times$ & 2.6$\times$ \\
MonoNet & 1.6$\times$ & 1.6$\times$ & 1.5$\times$ & 1.5$\times$ & 1.4$\times$ & 1.6$\times$ \\
\midrule
FedAvg (s per run) & 59 & 64 & 21 & 7 & 8 & 218 \\
\bottomrule
\end{tabular}

\end{table}

\subsection{Why did the server-side correction fail?}
\label{app:server}

Grad-Box + server takes five Adam steps (learning rate $10^{-3}$) on the box penalty after each aggregation, with a proximal term but no task loss. It increased violations relative to Grad-Box in 17 of 18 matched settings.
To check whether this is a property of server-side correction or of our implementation, we replayed German training (two seeds, $\lambda=10$) and measured the violated fraction on held-out box samples immediately before and after every server step (\texttt{scripts/diag\_server\_step.py}). The server step \emph{raised} the violated fraction in 100/100 rounds (seed 0) and 88/100 rounds (seed 1). The following client round \emph{lowered} it on average. A natural hypothesis, that clients undo the server's correction, is therefore wrong: the correction itself makes things worse.
The server step even increased its own objective, the box penalty, on held-out box points (100/100 rounds for seed 0). The increase persisted when we reduced Adam's learning rate to $10^{-4}$ and $10^{-5}$ (penalty up in 38--39 of 40 rounds), whereas plain gradient descent decreased the penalty (36 of 40 rounds), by a negligible amount.
Our explanation: Adam's coordinate-normalised first steps move \emph{every} parameter by about the learning rate, including parameters whose penalty gradient is negligible. Because the penalty is non-smooth in the parameters (ReLU activation patterns at the sampled points change), such steps create new violations elsewhere faster than they remove the few that drive the gradient.
The negative result is therefore an artefact of our choice of server optimiser, not evidence against server-side correction. A correctly tuned server step (e.g.\ gradient descent with a step size matched to the penalty scale, or a line search) might help. We did not rerun the sweep with such a step.

\end{document}